\documentclass[11pt]{article}

\usepackage{comment,url,graphicx,subcaption,relsize}
\usepackage{amssymb,amsfonts,amsmath,amsthm,amscd,dsfont,mathrsfs,mathtools,nicefrac}
\usepackage{float,psfrag,epsfig,color,xcolor,url,hyperref}
\usepackage{epstopdf,bbm,mathtools,enumitem}
\usepackage[toc,page]{appendix}
\usepackage[mathscr]{euscript}

\usepackage[top=1in, bottom=1in, left=1in, right=1in]{geometry}

\def\K{\mathcal{K}}

\def\balign#1\ealign{\begin{align}#1\end{align}}
\def\baligns#1\ealigns{\begin{align*}#1\end{align*}}
\def\balignat#1\ealign{\begin{alignat}#1\end{alignat}}
\def\balignats#1\ealigns{\begin{alignat*}#1\end{alignat*}}
\def\bitemize#1\eitemize{\begin{itemize}#1\end{itemize}}
\def\benumerate#1\eenumerate{\begin{enumerate}#1\end{enumerate}}

\newenvironment{talign*}
 {\csname align*\endcsname}
 {\endalign}
\newenvironment{talign}
 {\csname align\endcsname}
 {\endalign}

\def\balignst#1\ealignst{\begin{talign*}#1\end{talign*}}
\def\balignt#1\ealignt{\begin{talign}#1\end{talign}}

\let\originalleft\left
\let\originalright\right
\renewcommand{\left}{\mathopen{}\mathclose\bgroup\originalleft}
\renewcommand{\right}{\aftergroup\egroup\originalright}

\def\tinycitep*#1{{\tiny\citep*{#1}}}
\def\tinycitealt*#1{{\tiny\citealt*{#1}}}
\def\tinycite*#1{{\tiny\cite*{#1}}}
\def\smallcitep*#1{{\scriptsize\citep*{#1}}}
\def\smallcitealt*#1{{\scriptsize\citealt*{#1}}}
\def\smallcite*#1{{\scriptsize\cite*{#1}}}

\def\R{\mathbb{R}}

\def\<{\left\langle} %
\def\>{\right\rangle}

\newcommand{\normtwo}[1]{\ensuremath{\!|\!| #1 | \! |_{2}}}

\newcommand{\bigo}{\mathcal{O}}

\DeclareSymbolFont{rsfs}{U}{rsfs}{m}{n}
\DeclareSymbolFontAlphabet{\mathscrsfs}{rsfs}

\providecommand{\argmin}{\mathop\mathrm{arg min}}

\ifdefined\nonewproofenvironments\else
\ifdefined\ispres\else
\newtheorem{theorem}{Theorem}
\newtheorem{lemma}[theorem]{Lemma}
\newtheorem{corollary}[theorem]{Corollary}

\newtheorem{example}{Example}[section]

\renewenvironment{proof}{\noindent\textbf{Proof.}\hspace*{.3em}}{\qed\\}
\newenvironment{proof-sketch}{\noindent\textbf{Proof Sketch}
  \hspace*{1em}}{\qed\bigskip\\}
\newenvironment{proof-idea}{\noindent\textbf{Proof Idea}
  \hspace*{1em}}{\qed\bigskip\\}
\newenvironment{proof-of-lemma}[1][{}]{\noindent\textbf{Proof of Lemma {#1}}
  \hspace*{1em}}{\qed\\}
\newenvironment{proof-of-theorem}[1][{}]{\noindent\textbf{Proof of Theorem {#1}}
  \hspace*{1em}}{\qed\\}
\newenvironment{proof-attempt}{\noindent\textbf{Proof Attempt}
  \hspace*{1em}}{\qed\bigskip\\}

\fi

\newtheorem{proposition}[theorem]{Proposition}

\newtheorem{assumption}{Assumption}
\fi
\makeatletter
\@addtoreset{equation}{section}
\makeatother

\hypersetup{
  colorlinks,
  linkcolor={red!50!black},
  citecolor={blue!50!black},
  urlcolor={blue!80!black}
}

\newcommand{\eq}[1]{\begin{align}#1\end{align}}
\newcommand{\eqn}[1]{\begin{align*}#1\end{align*}}

\usepackage{algorithm}%
\usepackage{algorithmicx}%
\usepackage{algpseudocode}%
\usepackage{listings}%
\usepackage{tikz}
\usetikzlibrary{decorations.pathreplacing,calc}

\newcommand{\rp}{\mathbb{R}^p}

\newcommand{\ball}{\mathcal{B}}

\newcommand{\wstwo}{{\sf W}_2}
\newcommand{\wsq}{{\sf W}_q}
\newcommand{\wsone}{{\sf W}_1}

\newcommand{\proj}{\mathtt{P}}

\def\btheta{\boldsymbol\theta}

\def\bxi{\boldsymbol\xi}

\def\bfA{\mathbf A}
\def\bfB{\mathbf B}
\def\bB{\mathbf B}
\def\bG{\mathbf G}
\def\bv{\mathbf v}
\def\bV{\mathbf V}
\def\bfI{\mathbf I}
\def\bL{\boldsymbol L}
\def\bW{\boldsymbol W}

\def\wass{{\sf W}}

\def\bvartheta{\boldsymbol\vartheta}

\def\rmd{{\rm d}}

\usepackage{mathtools}

\newcommand{\1}{\mathds 1}

\usepackage[utf8]{inputenc} %
\usepackage[T1]{fontenc}    %
\usepackage{hyperref}       %
\usepackage{url}            %
\usepackage{booktabs}       %
\usepackage{amsfonts}       %
\usepackage{nicefrac}       %
\usepackage{microtype}      %
\usepackage{xcolor}         %
\usepackage{mathtools}
\usepackage{multirow}

\mathtoolsset{showonlyrefs}

\definecolor{darkmidnightblue}{rgb}{0.0, 0.2, 0.4}
\definecolor{darkpowderblue}{rgb}{0.0, 0.2, 0.6}
\definecolor{dukeblue}{rgb}{0.0, 0.0, 0.61}

\hypersetup{
    colorlinks = true,
    citecolor= midnightblue,
    urlcolor= black,
    breaklinks=true,
    linkcolor = dukeblue,
    linkbordercolor = {white},
}

\definecolor{darkmidnightblue}{HTML}{003366}    
\definecolor{midnightblue}{HTML}{0059b3}
\definecolor{chromered}{HTML}{f14233}

\newcommand{\logo}{\widetilde{\mathcal{O}}}

\begin{document}

\title{Randomized Midpoint Method for Log-Concave Sampling under Constraints}

 \author{
Yifeng Yu\thanks{Department of Mathematical Sciences, Tsinghua University \texttt{yyf22@mails.tsinghua.edu.cn}}
 \and 
 Shijie Zhang\thanks{Artificial Intelligence Research Institute, Shenzhen MSU-BIT University, \texttt{shijie.z@smbu.edu.cn}}
 \and
 Lu Yu\thanks{
Department of Data Science, City University of Hong Kong \texttt{lu.yu@cityu.edu.hk}
 }
}

\maketitle

\begin{abstract}
In this paper, we study the problem of sampling from log-concave distributions supported on convex and compact sets, with a particular focus on the randomized midpoint discretization of both overdamped and kinetic Langevin diffusions in constrained domains. We revisit the proximal framework for handling constraints through projection operators and develop a more general formulation that encompasses Euclidean, Bregman, and Gauge projections. The resulting smooth approximation allows a unified and tractable analysis of Langevin algorithms and their variants under constraints.
Within this framework, we establish convergence guarantees in Wasserstein-$q$ $(q\geqslant 1)$ distances between the smooth surrogate and the target distribution. 
We further derive complementary lower bounds, showing that the results are near-optimal in order. Building upon this tight approximation analysis, we obtain new convergence guarantees for the randomized midpoint Langevin algorithms and refined bounds for both vanilla and kinetic Langevin Monte Carlo methods under constraints, thereby advancing the theoretical understanding of constrained diffusion-based sampling.
\end{abstract}

\section{Introduction}

Sampling from probability distributions plays a critical role in various fields of science and engineering, especially when dealing with convex and compact sets~\cite{andrieu2003introduction,gelman1995bayesian,stuart2010inverse}. In this context, the problem involves sampling from a probability measure $\nu$ on such sets, characterized by its density function
\eqn{
\nu(\btheta)=\frac{e^{-U^{}(\btheta)}}{\int_{\mathbb{R}^p} e^{-U^{}(\btheta)}\rmd \btheta}\,,
}
Here, $U(\btheta) = f(\btheta) + \ell_{\mathcal{K}}(\btheta)$, where $f:\R^d\to\R$ represents a potential function and $\ell_{\mathcal{K}}(\btheta)$ is an indicator function ensuring $\btheta$ lies within the convex and compact set $\mathcal{K}\subset \mathbb{R}^p$
\eqn{
\ell_{\K}(\btheta):=
\begin{cases}
+\infty & \text{ if } \btheta\notin\K\\
 0& \text{ if } \btheta\in\K\,.
\end{cases}
}
Solving this constrained sampling problem is challenging and has garnered considerable interest across various fields, including computer science and statistics. 

In the absence of constraints, a wide range of diffusion-based samplers have been developed, leveraging discretizations of Itô diffusions to efficiently approximate the target distribution. A canonical example is Langevin Monte Carlo~\cite{Dalalyan14,durmus2017,erdogdu2021convergence,erdogdu2022convergence,Erdogdu18,mou2022improved,mousavi2023towards,raginsky2017non, RobertsTweedie96}, a Markov chain Monte Carlo method that simulates the motion of a particle in a potential defined by the target density. 
It corresponds to the Euler-Maruyama discretization of the Langevin diffusion.
A natural extension, kinetic Langevin Monte Carlo, discretizes the kinetic Langevin diffusion, which incorporates momentum and friction to balance exploration and exploitation~\cite{einstein1905molekularkinetischen,von1906kinetischen}. 
 As shown by~\cite{Nelson}, the Langevin diffusion is a rescaled limit of the kinetic version. Theoretical properties such as ergodicity and mixing time have been studied in works like~\cite{dalalyan_riou_2018,eberle2019}.  
To improve discretization accuracy,~\cite{shen2019randomized} propose the randomized midpoint scheme for kinetic Langevin diffusion, showing its advantages in error tolerance and condition number dependence. 
Subsequent work further analyzed its probabilistic properties~\cite{he2020ergodicity,yu2023langevin}, 
its suitability for parallel implementation~\cite{yu2024parallelized}, and its performance and acceleration benefits in generative modeling~\cite{gupta2024faster,li2024improved,yu2025advancing,kandasamy2024poisson}.
Despite these advances, the behavior of midpoint randomized Langevin methods in the presence of constraints remains largely unexplored. While the underlying continuous-time processes, such as those associated with the Langevin algorithms, can in principle be extended to constrained domains, existing discretization analyses typically rely on smoothness assumptions of the potential function. These assumptions fail when hard constraints are present, since the indicator function $\ell_{\mathcal{K}}$ introduces non-differentiability and discontinuities near the boundary of the feasible set. As a result, establishing convergence guarantees and stable numerical schemes for constrained Langevin sampling remains a challenging problem.

Earlier work on constrained sampling~\cite{brosse2017sampling,gurbuzbalaban2022penalized} addresses the non-smoothness of the target density $\nu$ by introducing smooth surrogate functions to approximate the indicator of the feasible set. In particular, these studies adopt a regularization approach that relies on projections onto $\mathcal{K}$, which effectively transforms the constrained sampling problem into an unconstrained one and enables a tractable analysis of diffusion-based algorithms.
This leads to the following error decomposition in Wasserstein-$q$ distance
\begin{align}
\wsq(\nu_n, \nu) \leqslant \underbrace{\wsq(\nu_n, \nu^\lambda)}_{\text{sampling error}} + \underbrace{\wsq(\nu^\lambda, \nu)}_{\text{approximation error}},\qquad q\in\{ 1,2\}\,,
\label{eq:err_decomp}
\end{align}
where $\lambda>0$ is the tuning parameter of the regularization, $\nu^\lambda$ denotes the surrogate distribution associated with the regularized potential, and $\nu_n$ represents the distribution of the sampler output after $n$ iterations. 
The first term captures the sampling error introduced by the algorithm, and the second represents the approximation bias resulting from the smoothing procedure.

Building on the foundations established by~\cite{brosse2017sampling,gurbuzbalaban2022penalized}, we extend this smoothing strategy to the randomized midpoint Langevin framework. 
In contrast to prior approaches that are typically designed around specific regularization schemes and primarily depend on the Euclidean projection, which corresponds to the Moreau envelope of the indicator function~$\ell_{\K}$~\cite{brosse2017sampling,durmus2018efficient,pereyra2016proximal,rockafellar2009variational}, our framework accommodates a broader class of projection operators. These include metric-based projections such as the Euclidean projection; more generally, Bregman projections, which are widely applied in clustering, classification, and outlier detection~\cite{ghorbani2019mahalanobis,xiang2008learning,xing2002distance}; and scaling-based projections, such as the Gauge projection~\cite{lu2022projection,mhammedi2022efficient}, which can be efficiently implemented using a membership oracle~\cite{mhammedi2022efficient}.
We show that the proposed smooth surrogate density provides a tight approximation to the true target density in Wasserstein-$q$ distance for any $q \geqslant 1$ by explicitly characterizing $\wsq(\nu^\lambda, \nu)$. 
When $q=1$, our bound recovers the rate in \cite{brosse2017sampling} while removing the need for convexity of $f$.
For $q=2$, we obtain a strictly sharper convergence rate than \cite{gurbuzbalaban2022penalized}, together with explicit constants and a transparent dependence on all relevant parameters. These improvements rely on a new proof technique based on a carefully constructed transport map that aligns the radial components of the surrogate distribution $\nu^\lambda$ with those of the target distribution $\nu$, enabling tight control of the Wasserstein distance.
To assess the optimality of these results, we further derive the first lower bound on the Wasserstein distance between the surrogate and the target, complementing the existing upper bounds.
Our refined approximation analysis also strengthens the convergence guarantees for the constrained (kinetic) Langevin Monte Carlo samplers (referred to as CLMC and CKLMC in this work) originally introduced in \cite{brosse2017sampling,gurbuzbalaban2022penalized}, leading to improved iteration complexity bounds.
Finally, we incorporate the randomized midpoint scheme into both vanilla and kinetic Langevin diffusions to sample from
$\nu^\lambda,$ resulting in the CRLMC and CRKLMC algorithms. By combining the sampling error of these schemes with our approximation error for $\nu^\lambda$, we obtain complete convergence guarantees for randomized midpoint Langevin Monte Carlo under the constrained settings. 
Our main contributions are as follows.
\begin{itemize}
   \setlength\itemsep{0em}
   \item In Section~\ref{sec:approx_target}, we demonstrate that our proposed framework can incorporate various projection options, such as Euclidean, Bregman, and Gauge projections.
   Notably, the latter two have received limited attention in the context of constrained sampling.
\item 
In Section~\ref{sec:general}, we analyze the approximation error between the proposed surrogate density and the true target distribution in terms of the Wasserstein-$q$ ($q \geqslant 1$) distance. Our results are consistent with those in~\cite{brosse2017sampling} when $q = 1$, while relaxing the convexity requirement on $f$. 
For $q = 2$, our analysis sharpens the existing $\wstwo$ bound established in~\cite{gurbuzbalaban2022penalized}, providing explicit constants and a clear dependence on parameters. 
To establish these upper bounds, we introduce a novel proof technique grounded in optimal transport theory, which exploits radial alignment in polar coordinates to tightly control the Wasserstein distance between $\nu^\lambda$ and $\nu$. 
\item Furthermore, we derive complementary lower bounds in Proposition~\ref{prop:lb}, demonstrating the near-optimality of our upper bounds.
To the best of our knowledge, such lower-bound results have not been explored in prior work.
    \item In Section~\ref{sec:rlmc} and Section~\ref{sec:rklmc}, we introduce the CRLMC and CRKLMC algorithms and provide convergence analyses under both Wasserstein-1 and Wasserstein-2 distances, offering new insight into the behavior of randomized midpoint Langevin dynamics in constrained settings.
 \item  In Section~\ref{sec:lmc} and Section~\ref{sec:klmc}, we revisit the CLMC and CKLMC algorithms. 
    Building on our refined analysis of the approximation error between $\nu^\lambda$ and $\nu$, we derive sharper non-asymptotic guarantees on their sampling error, improving upon existing results. 
    
\end{itemize}
We develop a systematic and unified analysis of constrained LMC variants, yielding strengthened upper Wasserstein bounds for LMC and its variants.
Our framework extends seamlessly to sampling methods beyond the LMC family.
In addition, our results provide explicit constants and a precise dependence on initialization and algorithmic parameters, offering a principled basis for selecting step sizes and tuning parameters, as well as for comparing convergence rates across different methods.
Table~\ref{tab:1} summarizes all of our iteration complexity results, along with existing results for comparison.

\subsection{Related Works} In the realm of computer science, a line of research initiated by~\cite{dyer1991random} explored polynomial-time algorithms for uniformly sampling convex bodies. 
This has been followed by seminal studies on the convergence properties of the Ball Walk and the Hit-and-Run algorithm toward uniform density on a convex body or, more broadly, to log-concave densities~\cite{kannan1997random,kook2024and,lovasz1999hit,lovasz1993random,lovasz2007geometry,smith1984efficient,jiang2024regularized}. 
Unlike the first-order sampling methods examined in this work, these algorithms require calls to zero-order oracles at each iteration. Additionally, the performance of these geometric random walks is influenced by the skewed shape of $\K$, necessitating pre-processing steps to enhance efficiency.

Another category of samplers addresses the geometric structure of convex constraints, including Riemannian Hamiltonian Monte Carlo~\cite{brubaker2012family,kook2022sampling,girolami2011riemann}, Gibbs sampling~\cite{gelfand1992bayesian}.
Furthermore, advances in deep learning and generative modeling, such as generative adversarial networks~\cite{goodfellow2014generative} and variational autoencoders~\cite{kingma2013auto}, have led to powerful sampling methods for complex, high-dimensional distributions, including those defined on a convex body~\cite{ortiz2022structured}.

In recent years, leveraging optimization techniques to facilitate sampling has become a widely adopted strategy. Recasting the sampling task as an optimization problem has led to a broad family of methods that integrate tools from optimization, stochastic processes, and convex analysis. 
Notably, projected Langevin algorithms~\cite{bubeck2015finite,bubeck2018sampling,lamperski2021projected,lehec2023langevin} apply projection onto the constraint set to ensure feasibility, drawing inspiration from projected gradient descent.
Proximal methods~\cite{brosse2017sampling,durmus2018efficient,salim2020primal} further exploit the analytic structure by using proximal operators to manage nonsmooth potentials and to separate complex components of the dynamics.
Another line of work, known as Mirror Langevin, is inspired by mirror descent and explores sampling in non-Euclidean geometries by leveraging a mirror map to sample directly from constrained distributions~\cite{ahn2021efficient,chewi2020exponential,hsieh2018mirrored,zhang2020wasserstein,li2022mirror,sato2025convergence,srinivasan2024fast}.
More recently, drawing from classical optimization techniques for saddle point problems, Primal-Dual Langevin methods have been proposed to handle sampling under expectation constraints via a saddle-point reformulation~\cite{chamon2024constrained}.
Beyond gradient-based schemes, particle-based algorithms~\cite{li2022sampling} employ optimization-style updates across interacting particles to approximate transport maps. 
Among these works, \cite{brosse2017sampling,gurbuzbalaban2022penalized} are the most closely aligned with the spirit of this study. 
We highlight the most relevant findings from both studies alongside our main results.
A detailed comparison and discussion are provided in Section~\ref{sec:comp}.\\

\noindent \textbf{Notation.}
Denote the $p$-dimensional Euclidean space by $\rp$.
We use $\btheta$ for deterministic vectors and 
$\bvartheta$ for random vectors.
$\bfI_p$ and $\mathbf 0_p$ denote the $p \times p$ identity and zero matrices, respectively.  
For symmetric matrices $\bfA$ 
and $\bfB$, we write $\bfA
\preccurlyeq\bfB$ (or $\bfB\succcurlyeq \bfA$) 
if $\bfB - \bfA$ is 
positive semi-definite. 
For a measurable function $f:\R^p\to\R$ and a set $\K\subset\R^p$, define the oscillation
$
\operatorname{osc}_{\K}(f)\;:=\;\sup_{x\in\K} f(x)\;-\;\inf_{x\in\K} f(x).
$For a twice differentiable function $f$, we denote its gradient and Hessian by $\nabla f$ and $\nabla^2 f$, respectively.
For any set $\K\subset\rp,$ we denote its complement by $\K^c$, and its volume by $\text{Vol}(\K).$ 
$\delta_{x}$ denotes the Dirac measure concentrated at the point $x$.
Given probability measures $\mu$ and $\nu$ on $(\rp,\mathcal{B}(\rp))$, the Wasserstein-$q$ distance is defined as
\[
\wass_q(\mu,\nu) = \big(\inf_{\varrho\in \Gamma(\mu,\nu)} \int_{\rp\times \rp}
\normtwo{\btheta -\btheta'}^q\,\rmd\varrho(\btheta,\btheta')\big)^{1/q}, q\geqslant 1\,,
\]
where the infimum is over all couplings $\varrho$ with marginals $\mu$ and $\nu$.

\begin{table}[!h]
\begin{center}
\caption{Iteration complexities of \{\text{CL, CKL, CRL, CRKL}\}MC algorithms for achieving $\varepsilon$-accuracy in $\mathsf{W}_1$ and $\mathsf{W}_2$ distances under different projection operators (with Euclidean projection as a special case of Bregman projection).  
This table also includes existing results for CLMC (also known as MYULA~\cite{brosse2017sampling} and PSGLD~\cite{gurbuzbalaban2022penalized}) and CKLMC (also referred to as PSGULMC~\cite{gurbuzbalaban2022penalized}),
Projected LMC algorithm~\cite{bubeck2018sampling}, and Mirrored Langevin algorithm~\cite{hsieh2018mirrored}.
}
\label{tab:1}
\vspace{0.2cm}
\begin{tabular}{c|c c c}
\toprule
Algorithm & Projection & Metric & Complexity  \\
\midrule
Projected LMC in~\cite{bubeck2018sampling} & /& TV & $\widetilde{\mathcal{O}}(\varepsilon^{-12})$ \\
\midrule
Mirrored Langevin in~\cite{hsieh2018mirrored} & / & $\wstwo$ & $\widetilde{\mathcal{O}}(\varepsilon^{-2})$ \\
\midrule
\multirow{2}{*}{MYULA } & \multirow{2}{*}{Euclidean} & $\mathsf{W}_1$ & $\widetilde{\mathcal{O}}(\varepsilon^{-6})$ \\
                      &                                 & TV & $\widetilde{\mathcal{O}}(\varepsilon^{-6})$ \\
\midrule
PSGLD in~\cite{gurbuzbalaban2022penalized} & Euclidean & $\mathsf{W}_2$ & $\widetilde{\mathcal{O}}(\varepsilon^{-18})$ \\
\midrule
\multirow{2}{*}{CLMC~(Theorem~\ref{thm:lmc})} & \multirow{2}{*}{Bregman, Gauge} & $\mathsf{W}_1$ & $\widetilde{\mathcal{O}}(\varepsilon^{-4})$ \\
                      &                                 & $\mathsf{W}_2$ & $\widetilde{\mathcal{O}}(\varepsilon^{-6+4/p})$ \\
\midrule
PSGULMC in~\cite{gurbuzbalaban2022penalized} & Euclidean & $\mathsf{W}_2$ & $\widetilde{\mathcal{O}}(\varepsilon^{-13})$ \\
\midrule
\multirow{2}{*}{CKLMC~(Theorem~\ref{thm:klmc})} & \multirow{2}{*}{Bregman, Gauge} & $\mathsf{W}_1$ & $\widetilde{\mathcal{O}}(\varepsilon^{-4})$ \\
                       &                                 & $\mathsf{W}_2$ & $\widetilde{\mathcal{O}}(\varepsilon^{-7+6/p})$ \\
\midrule
\multirow{2}{*}{CRLMC~(Theorem~\ref{thm:rlmc})} & \multirow{2}{*}{Bregman, Gauge} & $\mathsf{W}_1$ & $\widetilde{\mathcal{O}}(\varepsilon^{-10/3})$ \\
                       &                                 & $\mathsf{W}_2$ & $\widetilde{\mathcal{O}}(\varepsilon^{-6+16/(3p)})$ \\
\midrule
\multirow{2}{*}{CRKLMC~(Theorem~\ref{thm:rklmc})} & \multirow{2}{*}{Bregman, Gauge} & $\mathsf{W}_1$ & $\widetilde{\mathcal{O}}(\varepsilon^{-8/3})$ \\
                        &                                 & $\mathsf{W}_2$ & $\widetilde{\mathcal{O}}(\varepsilon^{-5+14/(3p)})$ \\
\bottomrule
\end{tabular}
\end{center}
\end{table}

\section{Main results}
\label{sec:main}

We begin by introducing a smooth approximation of the target density and deriving Wasserstein bounds on its distance to the target.

\subsection{Smooth Approximation for the Target Density}
\label{sec:approx_target}
The absence of smoothness in the target distribution $\nu$ presents significant challenges because sampling algorithms often depend on the smoothness properties of the target distribution to effectively explore the space and produce representative samples.
Motivated by this, we consider the approximation for $\ell_{\K}$ 
\begin{align*}
    \ell_{\K}^{\lambda}(\btheta):=\dfrac{1}{2\lambda^2}d_{\K}(\btheta),
\end{align*}
where $\lambda>0$ is the tuning parameter and
$d_{\K}:\rp\to\mathbb R_+$ is a distance function that quantifies the distance between $\btheta$ and the constraint set $\K$.
We will later introduce specific choices for $d_{\K}.$
Define 
\eqn{
U^\lambda(\btheta):=f(\btheta)+\ell_{\K}^\lambda(\btheta)\,,  
}
and the corresponding surrogate target density $\nu^\lambda$ 
\eq{
\label{eq:surrogate}
\nu^\lambda(\btheta):=\frac{e^{-U^{\lambda}(\btheta)}}{\int_{\rp} e^{-U^{\lambda}(\btheta')}\rmd \btheta'}\,.
}
We assume the convex and compact set $\K$ satisfies the following assumption.
\begin{assumption}\label{asm:radius}
Given constants $0<r<R<\infty$, it holds that $\ball_2(r)\subset \K\subset \ball_2(R)$, where
$\ball_2(r)$ denotes the Euclidean ball of radius $r$ centered at the origin.
\end{assumption}
This assumption has been made in the work of constrained sampling~\cite{bubeck2018sampling,brosse2017sampling,gurbuzbalaban2022penalized,lamperski2021projected}.
For clarity of presentation, we assume in Assumption~\ref{asm:radius} that the origin $\mathbf{0}\in\K,$ though our results remain valid without this condition.
We also assume the potential function $f$ satisfies the following standard assumption.
\begin{assumption}\label{asm:f}
The function $f: \mathbb R^p\to\mathbb R$ 
is continuous and is lower bounded on $\R^p$.
\end{assumption}
Furthermore, we require the distance function $d_{\K}$ to satisfy the following assumption.
\begin{assumption}
\label{asm:dK}
There exists constants $0<c_1\leqslant c_2$ such that $c_1\|\btheta-\proj_{\K}(\btheta)\|_2^{2}\leqslant d_{\K}(x)\leqslant c_2\|\btheta-\proj_{\K}(\btheta)\|_2^{2}$, where $\proj_{\K}:\R^p\to\R^p$ is a projection operator onto set $\K.$
Moreover, $d_{\K}(\btheta)\geqslant 0$ for all $\btheta\in\R^p$, and $d_{\K}(\btheta)=0$ whenever $\btheta\in\K.$
\end{assumption}
We note that the choice $d_{\K} = \|\btheta - \proj_{\K}(\btheta)\|_2^{2}$, where $\proj_{\K}$ denotes the Euclidean projection onto $\K$, satisfies this assumption. This formulation is also adopted in~\cite{gurbuzbalaban2022penalized} (Theorem 2.7) and~\cite{brosse2017sampling}.
In the following, we introduce two general approximation schemes for $\ell_{\mathcal{K}}$, corresponding to two broad classes of projections: \textbf{metric-based} and \textbf{scaling-based}. For each projection type, we specify the corresponding choice of $d_{\mathcal{K}}$ and show that both constructions satisfy Assumption~\ref{asm:dK}, providing flexible and effective smooth surrogates for the indicator function.

\begin{example}
\label{ex:Maha}
[Metric-based Projection]
Consider the Bregman Projection $\proj_{\K}^B:\R^p\to\K$
\eq{
\label{eq:Maha_fnc}
\proj_{\K}^B(\btheta):=\argmin_{\btheta'\in \K}(\btheta-\btheta')^\top Q(\btheta-\btheta')
}
where $Q\in\R^{d\times d}$ is a positive semi-definite symmetric matrix.
\end{example}
Note that the Bregman divergence $D_F(\btheta,\btheta'):=(\btheta-\btheta')^\top Q(\btheta-\btheta')$ is the squared Mahalanobis distance, generated by function $F(\btheta):=\frac{1}{2}\btheta^\top Q \btheta$. 
In this case, we consider the distance function of the form
\begin{align}\label{eq:disEuc}
    d_{\K}(\btheta)=\|\btheta-\proj_{\K}^{B}(\btheta)\|_2^2.
\end{align}
Therefore, the Bregman projection induces the following approximation to $\ell_{\K}$
\eqn{
\ell_{\K}^{B,\lambda}(\btheta)=\dfrac{1}{2\lambda^2}\|\btheta-\proj_{\K}^B(\btheta)\|_2^2=\inf_{\btheta'\in\R^p}\left(\ell_{\K}(\btheta')+\dfrac{1}{2\lambda^2}(\btheta-\btheta')^\top Q(\btheta-\btheta')\right).
}
and $U^{B,\lambda}(\btheta)=f(\btheta)+\ell_{\K}^{B,\lambda}(\btheta).$ When $Q=\mathbf{I}_p,$ the Bregman projection recovers the Euclidean projection, which yields the Moreau envelope.

\begin{example}
\label{ex:gauge}
[Scaling-based projection]
The Gauge projection $\proj_{\K}^G:\rp\to\K$ is defined as
\eqn{
\proj_{\K}^G(\btheta):=\frac{\btheta}{g_{\K}(\btheta)}
}
where  $g_{\K}:\rp\to\R$ is a variant of the Gauge function (also known as the Minkowski functional) associated with the set $\K$, given by  
\eq{
\label{eq:gauge_fnc}
g_{\K}(\btheta):=\inf\{t\geqslant 1: \btheta\in t\K\}\,.
}
\end{example}
The function $g_{\K}$ characterizes the scaling required to project $\btheta$ onto the set $\K.$
In this case, we consider the distance function as
\eqn{
    d_{\K}(\btheta)=(g_{\K}(\btheta)-1)^2\,.
}

Unlike the metric-based projection~\eqref{eq:disEuc}, which inherits convexity from the Moreau envelope, the Gauge projection is defined within a non-Euclidean geometry. 
Moreover, under the assumption on $\K$, the distance function $d_{\K}$ satisfies Assumption~\ref{asm:dK} with constants $c_1=1/R^2$ and $c_2=1/r^2$.
Set $\ell_{\K}^{G,\lambda}(\btheta)=\frac{1}{2\lambda^2}(g_{\K}(\btheta)-   1)^2$.
The corresponding surrogate function is defined via $U^{G,\lambda}(\btheta):=f(\btheta)+\ell_{\K}^{G,\lambda}(\btheta).$
\subsection{Wasserstein Distance Bounds Between Distributions $\nu^\lambda$ and $\nu$}
\label{sec:general}
In this section, 
we establish upper and lower Wasserstein bounds between $\nu$ and its approximation $\nu^{\lambda}$,  demonstrating that the distance vanishes as $\lambda$ approaches zero.
\begin{proposition}
\label{prop:w2_tight}
Under Assumptions~\ref{asm:radius}-\ref{asm:dK},
for any $q\geqslant 1$ and any
\[
\lambda \in \Bigl(0,\; \tfrac{\sqrt{c_1}\,r}{p+q} \wedge \tfrac{\sqrt{c_1}\,r\,e^{\operatorname{osc}_{\K}(f)}}{3\sqrt{\pi}\,\operatorname{Vol}(\K)\,p}\Bigr),
\]
it holds that
\begin{align*}
    \wsq(\nu,\nu^\lambda)\leqslant C(p,q)\cdot\begin{cases}
        \lambda, &1/p+1/q>1\\
        \lambda\log^{1/q}(\frac{1}{\lambda}), &1/p+1/q=1\\
        \lambda^{1/p+1/q}, &1/p+1/q<1\,.
    \end{cases}
\end{align*}
where $C(p,q)=C_0\left[\frac{e^{4\operatorname{osc}_{\K}(f)}}{c_1^{1/2}}\cdot\left(\frac{\max(R,1)}{\min(r,1)}\right)^{(2p+1)}\cdot p\right]^{1/p+1/q}$ and $C_0>0$ is a universal numeric constant.

\end{proposition}

\noindent 
This theorem captures a phase transition in $\wsq(\nu,\nu^\lambda)$ that depends on the relation between dimension $p$ and Wasserstein order $q,$ exhibiting linear convergence when $1/p+1/q>1$, and slower rates in higher dimensions or for higher-order distances.
When $q=1,$ our result aligns with the bound $\wsone(\nu,\nu^\lambda)=\bigo(\lambda)$ stated in~\cite[Proposition 5]{brosse2017sampling}.
Moreover, our analysis relaxes the convexity assumption on $f$ required in~\cite{brosse2017sampling}.
For $q=2,$ our analysis yields the rate 
$\wass_2(\nu,\nu^\lambda)=\bigo(\lambda^{1/2+1/p})$ when $p>2$.
In comparison, \cite{gurbuzbalaban2022penalized} studies a special case of our Example 2.1 ($Q=I$), obtaining the bound {$\wstwo(\nu,\nu^\lambda)=\bigo(\lambda^{1/4}\log^{1/8}(1/\lambda^2))$}. 
Thus, our result provides a substantially sharper bound.

Our improved bounds follow from a new optimal transport argument that constructs a radial map $T$ aligning $\nu^\lambda$ with the target $\nu.$
Using the decomposition
\[
\wsq(\nu,\nu^\lambda)\leqslant \wsq(\nu,T_{\#}\nu^\lambda)+\wsq(T_{\#}\nu^\lambda,\nu^\lambda)\,,
\]
we isolate the small region near the origin where the two distributions differ and bound the transport cost in each radial regime.
Since $T$ aligns the distributions perfectly outside this small region, we avoid relying on coarse global bounds and obtain sharper non-asymptotic bounds.

To assess the optimality of our upper bounds, we also derive the following lower bound on $\wsq(\nu^\lambda,\nu)$.

\begin{proposition}\label{prop:lb}
Under Assumptions~\ref{asm:radius}-\ref{asm:dK}, for any $q\geqslant 1$, if 
\begin{align}
\label{cd:lb}
    \dfrac{\int_{\K}\normtwo{\btheta}^qe^{-U(\btheta)}\rmd \btheta}{\int_{\K}e^{-U(\btheta)}\rmd \btheta}\neq \lim_{\lambda\to 0}\dfrac{\int_{\K^c}\normtwo{\btheta}^qe^{-U^{\lambda}(\btheta)}\rmd \btheta}{\int_{\K^c}e^{-U^{\lambda}(\btheta)}\rmd \btheta}\,,
\end{align}
and if $f$ and $d_{\K}$ are both smooth, then for a sufficiently small $\lambda$, it holds that
$\wsq(\nu,\nu^{\lambda})=\Omega(\lambda)\,.$
\end{proposition}
Condition~\eqref{cd:lb} states that the conditional $q$-th moment over $\K^c$ under the surrogate density does not converge to that of the target distribution supported on $\K$ as $\lambda\to0.$
This condition is mild. For example, if $\K$ is a ball of radius $R$, it then follows that
\begin{align*}
\dfrac{\int_{\mathcal K^c}\|\btheta\|_2^qe^{-U^\lambda(\btheta)}d\btheta}{\int_{\mathcal K^c}e^{-U^\lambda(\btheta)}d\btheta}>R^q\,.
\end{align*}
Meanwhile, the smoothness of $U$ on $\K$ implies that the conditional $q$-th moment under $\nu$ is strictly smaller than $R^q$; namely, there exists $\delta>0$ such that
\begin{align*}
\dfrac{\int_{\mathcal K}\|\btheta\|_2^qe^{-U(\btheta)}d\btheta}{\int_{\mathcal K}e^{-U(\btheta)}d\btheta}\leqslant R^q-\delta\,.
\end{align*}
Hence, the two quantities do not coincide as $\lambda\to 0$, and Condition~\eqref{cd:lb} holds.

To our knowledge, this represents the first lower bounds for the Wasserstein distance between the surrogate distribution and the target distribution. 
Based on the results outlined in the preceding proposition, when $1/p+1/q>1$ and with a sufficiently small $\lambda,$ the lower bound aligns with the rate estimates from Proposition~\ref{prop:w2_tight} up to a constant factor, thereby confirming the optimality of the upper bound for this case.

\subsection{Randomized Midpoint Method for Langevin Diffusion under Constraints}
\label{sec:rlmc}
Having constructed the smooth surrogate density $\nu^\lambda$, which closely approximates the target distribution, our next objective is to sample from it.
To this end, we propose using Langevin algorithms tailored to the constrained setting of this work.

Throughout this section, we assume that the potential $f$ and the distance function $d_{\K}$ satisfy the following assumption.
\begin{assumption}
\label{asm:f1}
The distance function $d_{\K}$ is $M_0$-smooth for a constant $0<M_0<\infty$. Moreover, the function $f:\R^p\to \R$ is twice differentiable with Hessian satisfying $m \preccurlyeq  \nabla^2 f \preccurlyeq M$
for some constants $0<m\leqslant M<\infty.$
\end{assumption}
{
Under this assumption, the surrogate potential $U^\lambda$ is $m$-strongly convex and $M^\lambda$-smooth, with
$M^\lambda = M + M_0/\lambda^{2}$. Moreover, together with Assumptions~\ref{asm:radius} and~\ref{asm:f}, the constructions in
Examples~\ref{ex:Maha} and~\ref{ex:gauge} imply that $U^\lambda$ is strongly convex, smooth, and integrable.
}
\begin{lemma}
\label{lem:scv-smooth}
    Under Assumption~\ref{asm:f1}, the surrogate potential $U^{\lambda}$ is $m$-strongly convex and $M^\lambda$-smooth with $M^\lambda=M+M_0/\lambda^2$, where $M_0$ depends on both the geometry of the constraint set and the choice of projection as follows
    \begin{itemize}
        \item  Bregman Projection (Example~\ref{ex:Maha}): $M_0=\lambda_{\max}(Q)$, where $\lambda_{\max}$ is the largest  eigenvalue of the matrix $Q.$
        \item  Gauge Projection (Example~\ref{ex:gauge}): If the boundary $\partial\K$ is globally $C^2$-smooth, and the Hessian of the Gauge function is non-degenerate, then $M_0=C_{\K} r$, where $C_\K$ depends on the maximum of the principle curvatures along $\partial \K$.
    \end{itemize}
     Furthermore, if Assumptions~\ref{asm:radius} and~\ref{asm:f} also hold, then the potential is integrable
    \[
    \int_{\rp}e^{-U^\lambda}(\btheta)\,\rmd \btheta<\infty.
    \]
\end{lemma}
These favorable properties enable the application of Langevin Monte Carlo and its variants, originally designed for unconstrained settings, to sample from $\nu^\lambda\propto\exp(-U^\lambda).$
Let $\bvartheta_0\in\R^p$ be a random vector drawn from a distribution $\nu_0$ and let $\bW 
=(\bW_t: t\geqslant 0)$ be a $p$-dimensional 
Brownian motion that is independent of $\bvartheta_0$. 
One can sample from the approximate distribution $\nu^{\lambda}$ using vanilla Langevin diffusion, defined as the strong solution to the stochastic differential equation (SDE)
\eq{
 \label{eq:LD}
    \rmd\bL_t^{\sf LD} = -\nabla U^{\lambda}(\bL_t^{\sf LD})\, \rmd t 
    + \sqrt{2}\,\rmd\bW_t,
    \qquad t\geqslant 0,\qquad \bL^{\sf LD}_0=\bvartheta_0.
}
This equation has a unique strong solution, which 
is a continuous-time Markov process, termed Langevin 
diffusion. 
Under the further assumptions on the potential~$U^{\lambda}$, such as strong convexity, the Langevin diffusion is ergodic, geometrically mixing and has $\nu^\lambda$ as its unique invariant  distribution~\cite{bhattacharya1978}. 
Moreover, $\nu^\lambda$ can be sampled via a suitable discretization of Langevin diffusion, as done in the Langevin Monte Carlo algorithm using Euler discretization.
Specifically, for a small step size
$h > 0$ and $\Delta_h\bW_t = \bW_{t+h} - 
\bW_t$, the following approximation holds
\eqn{
    \bL^{\sf LD}_{t+h} & = \bL^{\sf LD}_t- \int_0^h  
    \nabla U^{\lambda}(\bL_{t+s}^{\sf LD})\,\rmd s + \sqrt{2}\;\Delta_h\bW_t \approx \bL_t^{\sf LD} - h  
    \nabla U^{\lambda}(\bL_t^{\sf LD}) + \sqrt{2}\;\Delta_{h}
    \bW_t. 
}
Motivated by this,~\cite{brosse2017sampling,gurbuzbalaban2022penalized} propose to use the Markov chain $(\bvartheta^{\sf CLMC}_k)_{k\in\mathbb N_+}$, which converges to $\nu^\lambda$ as the step size $h$ goes to zero. 
Approximating $\bvartheta^{\sf CLMC}_k\approx \bL^{
\sf LD}_{kh}$, the update rule is
\eq{
\label{eq:clmc}
    \bvartheta_{k+1}^{\sf CLMC} = \bvartheta_k^{\sf CLMC} 
    -h\nabla U^{\lambda}(\bvartheta_k^{\sf CLMC})
    +\sqrt{2}\,(\bW_{(k+1)h}-\bW_{kh}).
}
As an alternative to the Euler discretization method for SDE~\eqref{eq:LD}, we consider the randomized midpoint method introduced in~\cite{shen2019randomized}.
Unlike the Euler method, the randomized midpoint method evaluates the integral $\int_0^h\nabla U^\lambda(\bL_{t+s}^{\sf LD})\,\rmd s$ at a random point within the time interval $[0,h]$ rather than at the start.
Let $\iota$ be a uniform random variable on 
$[0,1]$, independent of the Brownian motion $\bW$. 
The randomized midpoint method exploits the approximation
\eqn{
\bL^{\sf LD}_{t+h}=\bL^{\sf LD}_t - \int_0^h 
    \nabla U^\lambda(\bL^{\sf LD}_{t+s})\, \rmd s + \sqrt{2}\;\Delta_h\bW_t \approx \bL^{\sf LD}_t 
    - h \nabla U^{\lambda}(\bL^{\sf LD}_{t+h\iota})+
    \sqrt{2}\;\Delta_h\bW_t.
}
The constrained randomized midpoint Langevin Monte Carlo (CRLMC) algorithm is formally defined as follows for each iteration $k = 1, 2, \dots$

\noindent\textbf{Step 1}
Generate independent $\bxi'_k,\bxi''_k\sim \mathcal{N}(\mathbf{0},\bfI_p)$ and uniform random variable $\iota_k\sim\text{Unif}\,[0,1]$.\\
\textbf{Step 2}    Set $\bxi_k = \sqrt{\iota_k}\,\bxi_k' + 
    \sqrt{1-\iota_k}\bxi''_k$ and define the 
    $(k+1)$-th iterate $\bvartheta^{\sf CRLMC}_{k+1}$ by 
    \begin{align*} 
        \bvartheta_{k+\iota}^{\sf CRLMC} &=  
        \bvartheta_k^{\sf CRLMC} - h\iota_k \nabla
        U^{\lambda}(\bvartheta_k^{\sf CRLMC}) + \sqrt{2h \iota_k}
        \, \bxi'_{k} \\
        \bvartheta_{k+1}^{\sf CRLMC} & = 
        \bvartheta_k^{\sf CRLMC}  - h\nabla U^{\lambda} 
        (\bvartheta_{k+\iota}^{\sf CRLMC}) + \sqrt{2h}
        \, \bxi_{k}.
    \end{align*}
For sufficiently small step size $h$ and large $n$, the law of $\bvartheta^{\sf CRLMC}_n$ approximates $\nu^\lambda$. Together with the analysis of $\wsq(\nu,\nu^\lambda)$ in Proposition~\ref{prop:w2_tight} from Section~\ref{sec:general}, this leads to the following complete convergence guarantee for CRLMC.
{\begin{theorem}\label{thm:rlmc}
Under Assumptions~\ref{asm:radius}-~\ref{asm:f1}, choose $\lambda \in \Bigl(0,\; \tfrac{\sqrt{c_1}\,r}{p+q} \wedge \tfrac{\sqrt{c_1}\,r\,e^{\operatorname{osc}_{\K}(f)}}{3\sqrt{\pi}\,\operatorname{Vol}(\K)\,p}\Bigr)$. Let the step size $h$ be such that $M^\lambda h +
\kappa (M^\lambda h)^{3/2}\leqslant 1/4$ with $\kappa=M^\lambda/m.$
Then, for any $p>2$ and every $n\geqslant 1$, the distribution $\nu^{\sf CRLMC}_n$ of $\bvartheta^{\sf CRLMC}_n$ satisfies
\begin{align*}
    \wstwo(\nu_n^{\sf CRLMC},\nu)\leqslant 1.1e^{-\frac{mnh}{2}}\wstwo(\nu_0^{\sf CRLMC},\nu)+(2.4\sqrt{\kappa M^{\lambda}h}+1.77)M^{\lambda}h\sqrt{\frac{p}{m}}
    +2.1C(p,2)\lambda^{\frac{1}{p}+\frac{1}{2}}\,.
\end{align*}
Moreover, if the initial point $\bvartheta_0^{\sf CRLMC}$ is at the minimizer of the function $U^\lambda$, then 
\begin{align*}
\wsone(\nu_n^{\sf CRLMC},\nu) 
& \leqslant 
1.1e^{-\frac{mnh}{2}}\sqrt{\frac{p}{m}}
+(2.4\sqrt{\kappa M^{\lambda}h}+1.77)M^{\lambda}h\sqrt{\frac{p}{m}}
+C(p,1)\lambda \,.
\end{align*}
\end{theorem}
Adopting $M^\lambda = M + M_0/\lambda^2$ induced by the projection constructions in Examples~\ref{ex:Maha} and~\ref{ex:gauge} (Lemma~\ref{lem:scv-smooth}), the preceding theorem implies the following corollary.}
\begin{corollary}\label{cor:rlmc}
Let the target precision level $\varepsilon\in(0,1)$ be small.\\
\noindent\textbf{(a)} Set $\lambda=\Theta(h^{3p/(9p+2)})$, and choose step size $h>0$ and $n\in\mathbb N_+$ so that
\begin{align*}
    h=\mathcal O\left(\varepsilon^{\frac{18p+4}{3p+6}}\right)\quad \text{and}\quad n=\widetilde{\Omega}\left(\varepsilon^{-\frac{18p+4}{3p+6}}\right)\,,
\end{align*}
then we have $\wstwo(\nu_n^{\sf CRLMC})=\logo(\varepsilon).$ \\
\noindent\textbf{(b)} Set $\lambda=\Theta(h^{3/10})$, and choose $h>0$ and $n\in\mathbb N_+$ so that
\begin{align*}
    h=\mathcal O\left(\varepsilon^{\frac{10}{3}}\right)\quad \text{and}\quad  n=\widetilde{\Omega}\left(\varepsilon^{-\frac{10}{3}}\right)\,,
\end{align*}
then we have $\wsone(\nu_n^{\sf CRLMC})=\logo(\varepsilon).$\footnote{The notation $\widetilde{\bigo}$ and $\widetilde{\Omega}$ suppress the dependence on $\log(1/\varepsilon)$}
\end{corollary}
We observe that for $p\geqslant 2$, it holds that $\frac{18p+4}{3p+6}\geqslant 6-\frac{16}{3p}$. Hence, the requirement on $n$ to achieve $\wstwo(\nu_n^{\sf CRLMC})=\logo(\varepsilon)$ can be simplified to $n=\widetilde{\Omega}(\varepsilon^{-6+16/(3p)}).$

\subsection{Randomized Midpoint Method for Kinetic Langevin Diffusion under Constraints}
\label{sec:rklmc}

The randomized midpoint method, introduced in~\cite{shen2019randomized}, provides a discretization of the kinetic Langevin process that reduces sampling bias compared to standard schemes in the unconstrained setting. 
We now examine its convergence behavior in the constrained framework considered here.
Recall that the kinetic Langevin process $\bL^{\textup{\sf KLD}}$
is a solution to the second-order SDE
\eq{\label{KLD:1}
{\textstyle\frac1{\gamma}}\ddot\bL_t^{\sf KLD} + 
  \dot\bL_t^{\sf KLD} = -\nabla U^{\lambda}(\bL_t^{\sf KLD}) 
  + \sqrt{2}\,\dot\bW_t,
}
with initial conditions $\bL_0^{\sf KLD} = 
\bvartheta_0$ and $\dot\bL_0^{\sf KLD} = \bv_0$, where $\gamma>0$ is the friction parameter and $\bW$ is a standard $p$-dimensional Brownian motion. Dots denote derivatives with respect to time $t \geqslant 0$. 
This can be formalized using It\^o's calculus and introducing the velocity field 
$\bV^{\sf KLD}$ so that the joint process 
$(\bL^{\sf KLD}, \bV^{\sf KLD})$ satisfies
\eq{\label{KLD:2}
    \rmd\bL^{\sf KLD}_t = \bV_t^{\sf KLD}\,\rmd t;
    \quad  
    \tfrac1{\gamma}\rmd\bV^{\sf KLD}_t = -\big(\bV_t^{\sf KLD} + 
    \nabla U^{\lambda}(\bL_t^{\sf KLD})\big)\,\rmd t + 
    \sqrt{2}\, \rmd \bW_t.
}
Similar to the vanilla Langevin diffusion \eqref{eq:LD}, this Markov process is ergodic when the potential $U^{\lambda}$ is strongly 
convex~\cite{eberle2019}, and admits the invariant density
\eqn{
    p_*(\btheta_,\bv) \propto \exp\big\{-U^{\lambda}(\btheta) - {\textstyle\frac1{2\gamma}}\|\bv\|^2\big\}, \qquad
    \text{for all}\quad \btheta,\bv\in\mathbb R^p.
}
whose marginal in $\btheta$ recovers the target distribution $\nu^\lambda.$
Notably, the distribution of $\bL^{\sf KLD}$ approaches that of the 
vanilla Langevin process $\bL^{\sf LD}$ as $\gamma$ approaches infinity (see e.g. \cite{Nelson}). 

To discretize this continuous-time process for sampling from $\nu^\lambda$, we apply the midpoint randomization scheme to the kinetic Langevin diffusion. 
The resulting algorithm, termed CRKLMC,  is applied at each iteration $k=1,2,\ldots$

\noindent\textbf{Step 1}:  sample a uniform random variable $\iota_k\sim \text{Unif}[0,1]$ and random vectors $(\bxi_k',\bxi''_k,\bxi_k''')$ such that, conditional on $\iota_k=u$, their joint distribution matches  $\big(\bB_u - e^{-\gamma h u} \bG_u,
                \bB_1 -e^{-\gamma h}\bG_1,
                \gamma e^{-\gamma h }\bG_1
            \big)$,
        where $\bB$ is a $p$-dimensional Brownian motion
        and $\bG_t = \int_0^{t} e^{\gamma h s}\,\rmd 
        \bB_s$.\\
\textbf{Step 2}:  let $\psi(x) = (1-e^{-x})/x$ and define the 
    $(k+1)$-th iterate of CRKLMC by
    \begin{align*}
        \bvartheta_{k+\iota}^{\sf CRKLMC}&=\bvartheta_k^{\sf CRKLMC}+  
        \iota_k h \psi(\gamma h \iota_k ) \bv_k^{\sf CRKLMC} - {\iota_k h} \big(1 - \psi 
        (\gamma h \iota_k)\big)\nabla U^\lambda(\bvartheta_k^{\sf CRKLMC}) + 
        \sqrt{2h}\,\bxi'_k\\
        \bvartheta_{k+1}^{\sf CRKLMC} &= \bvartheta_k^{\sf CRKLMC} +  h
        \psi(\gamma h)\bv_k^{\sf CRKLMC} -  \gamma h^2( 1 - \iota_k)
        \psi\big(\gamma h(1-\iota_k)\big) \nabla U^\lambda 
        (\bvartheta_{k+\iota}^{\sf CRKLMC}) + \sqrt{2h} \bxi''_k\\
        \bv_{k+1}^{\sf CRKLMC} &= e^{-\gamma h}\bv_k^{\sf CRKLMC} - {\gamma} h
        e^{- \gamma h(1 -\iota_k)} \nabla U^\lambda(\bvartheta_{k+\iota}^{\sf CRKLMC}) 
        +  \sqrt{2h}\,\bxi'''_k.
    \end{align*}
Prior work~\cite{shen2019randomized,yu2023langevin} has demonstrated that the sequence $(\bv_n^{\sf CRKLMC},\bvartheta_n^{\sf CRKLMC})$ efficiently approximates $(\bV_{nh}^{\sf CRKLMC},\bL_{nh}^{\sf CRKLMC})$ with large $n$ and small step size $h$.
We now state the main result of this paper, which provides an upper bound on the error of the CRKLMC algorithm.
{
\begin{theorem}\label{thm:rklmc}
Under Assumptions~\ref{asm:radius}-~\ref{asm:f1}, choose $\lambda \in \Bigl(0,\; \tfrac{\sqrt{c_1}\,r}{p+q} \wedge \tfrac{\sqrt{c_1}\,r\,e^{\operatorname{osc}_{\K}(f)}}{3\sqrt{\pi}\,\operatorname{Vol}(\K)\,p}\Bigr)$. 
Set $\gamma$ and $h$ so that $\gamma\geqslant 5M^\lambda$ and $\gamma h\leqslant 0.1\kappa^{-1/6},$ where $\kappa=M^\lambda/m.$ 
Assume that the initial point $\bvartheta_0^{\sf CRKLMC}$ is the minimizer of the function $U^\lambda$, and that $\bv_0^{\sf CRKLMC}\sim\mathcal N_p(0,\gamma \bfI_p)$.
Then, for any $p>2$ and every $n\geqslant 1$, the distribution $\nu^{\sf CRKLMC}_n$ of $\bvartheta^{\sf CRKLMC}_n$ satisfies
\begin{align*}
  \wstwo(\nu_n^{\sf CRKLMC},\nu)\leqslant 1.6e^{-mnh}\wstwo(\nu_0^{\sf CRKLMC},\nu)+(\gamma h)^3\sqrt{\frac{\kappa p}{m}}+10(\gamma h)^{3/2}\sqrt{\frac{p}{m}}+3C(p,2)\lambda^{\frac{1}{p}+\frac{1}{2}}\,.
\end{align*}
Moreover, it holds that
\begin{align*}
\wsone(\nu_n^{\sf CRKLMC},\nu) 
& \leqslant 
1.1e^{-\frac{mnh}{2}}\sqrt{\frac{p}{m}}
+ (2.4\sqrt{\kappa M^{\lambda}h}+1.77)M^{\lambda}h\sqrt{\frac{p}{m}}
+C(p,1)\lambda \,.
\end{align*}
\end{theorem}
}
As specified in Examples~\ref{ex:Maha} and~\ref{ex:gauge}, we take {$M^\lambda=M+M_0/\lambda^2$}. Under this choice, the preceding theorem yields the following consequence.
\begin{corollary}
\label{cor:rklmc}
Let the error level $\varepsilon\in(0,1)$ be small and $\gamma=5(M+M_0/\lambda^2).$ \\
\noindent\textbf{(a)} Set $\lambda=\Theta(h^{6p/(15p+2)})$, and choose $h>0$ and $n\in\mathbb N_+$ so that
\begin{align*}
    h=\mathcal O\left(\varepsilon^{\frac{15p+2}{3p+6}}\right)\quad\text{and}\quad n=\widetilde{\Omega}\left(\varepsilon^{-\frac{15p+2}{3p+6}}\right)\,,
\end{align*}
then we have $\wstwo(\nu_n^{\sf CRKLMC})=\logo(\varepsilon).$\\
\noindent\textbf{(b)} Set $\lambda=\Theta(h^{3/8})$, and choose $h>0$ and $n\in\mathbb N_+$ so that
\begin{align*}
    h=\mathcal O\left(\varepsilon^{\frac{8}{3}}\right) \quad \text{and}\quad n=\widetilde{\Omega}\left(\varepsilon^{-\frac{8}{3}}\right)\,,
\end{align*}
then we have $\wsone(\nu_n^{\sf CRKLMC})=\logo(\varepsilon).$
\end{corollary}
When $p\geqslant 2$, it holds that $\frac{15p+2}{3p+6}\geqslant 5-\frac{14}{3p}$. 
This implies the requirement on the number of iterations $n$ for achieving $\wstwo(\nu_n^{\sf CRKLMC})=\logo(\varepsilon)$ can be relaxed to $n=\tilde{\Omega}(\varepsilon^{-5+14/(3p)}).$

\subsection{Improved Error Bounds for Constrained Langevin Monte Carlo}
\label{sec:lmc}

In this section, we revisit Langevin Monte Carlo algorithms in the constrained setting, referred to as CLMC in this work. The algorithm, described in~\eqref{eq:clmc}, has been studied in prior literature~\cite{gurbuzbalaban2022penalized,brosse2017sampling}. 
Building on the refined analysis of $\wsq(\nu, \nu^\lambda)$ from Section~\ref{sec:general}, we establish improved upper bounds for the sampling error of CLMC under constraints. The theorem below presents the resulting enhanced Wasserstein convergence rate.
{
\begin{theorem}
\label{thm:lmc}
Under Assumptions~\ref{asm:radius}-~\ref{asm:f1}, choose $\lambda \in \Bigl(0,\; \tfrac{\sqrt{c_1}\,r}{p+q} \wedge \tfrac{\sqrt{c_1}\,r\,e^{\operatorname{osc}_{\K}(f)}}{3\sqrt{\pi}\,\operatorname{Vol}(\K)\,p}\Bigr)$. 
Let the step size $h$ be such that $M^\lambda h \leqslant 1$.
Then, for any $p>2$ and every $n\geqslant 1$, the distribution $\nu^{\sf CLMC}_n$ of $\bvartheta^{\sf CLMC}_n$ satisfies
\begin{align*}
    \wstwo(\nu_n^{\sf CLMC},\nu)\leqslant 
    e^{-mnh}\wstwo(\nu_0^{\sf CLMC},\nu)
    + \sqrt{\frac{2M^\lambda ph}{m}}
    +2C(p,2)\lambda^{\frac{1}{p}+\frac{1}{2}}\,.
\end{align*}
Moreover, if the initial point $\bvartheta_0^{\sf CLMC}$ is at the minimizer of the function $U^\lambda$, then 
\begin{align*}
\wsone(\nu_n^{\sf CLMC},\nu) 
& \leqslant 
e^{-mnh}\sqrt{\frac{p}{m}}
+ \sqrt{\frac{2M^\lambda ph}{m}}
+C(p,1)\lambda \,.
\end{align*}
\end{theorem}
}
Before discussing related work, we record the following corollary of the preceding theorem obtained by setting $M^\lambda = M + M_0/\lambda^{2}$.
\begin{corollary}
    \label{cor:lmc}
Let the error level $\varepsilon\in(0,1)$ be small.\\
\noindent\textbf{(a)} Set
$\lambda=\Theta\left(h^{p/(3p+2)}\right),$
and choose $h>0$ and $n\in\mathbb N_+$ so that
    \begin{align*}
        h=\mathcal{O}\left(\varepsilon^{\frac{6p+4}{p+2}}\right)\quad\text{and}\quad n=\widetilde{\Omega}\left(\varepsilon^{-\frac{6p+4}{p+2}}\right)\,,
    \end{align*}
    then we have $\wstwo(\nu_n^{\sf CLMC},\nu)=\logo(\varepsilon).$\\
\noindent\textbf{(b)} Set $\lambda=\Theta(h^{1/4})$, choose $h>0$ and $n\in\mathbb N_+$ so that
\begin{align*}
    h=\mathcal O\left(\varepsilon^4\right)\quad\text{and}\quad n=\widetilde{\Omega}\left(\varepsilon^{-4}\right)\,,
\end{align*}
then we have $\wsone(\nu_n^{\sf CLMC},\nu)=\logo(\varepsilon).$
\end{corollary}

According to this theorem, the CLMC algorithm requires $\tilde\bigo(\varepsilon^{-4})$ gradient evaluations to achieve a target accuracy $\varepsilon$ in $\wsone$ distance.
This improves upon the rate $\tilde\bigo(\varepsilon^{-6})$ obtained in~\cite{brosse2017sampling}.
Moreover, we note that $\frac{6p+4}{p+2}\geqslant 6-\frac{4}{p}$ holds when $p\geqslant 2$. 
Thus, the requirement on $n$ to achieve $\wstwo(\nu_n^{\sf CLMC},\nu)=\logo(\varepsilon)$ can be simplified to $n=\widetilde{\Omega}(\varepsilon^{-6+4/p}).$
This rate $\tilde\bigo(\varepsilon^{-6+4/p})$, significantly improves upon the result $\logo(\varepsilon^{-18})$ from Proposition 2.21 of~\cite{gurbuzbalaban2022penalized}, obtained under the zero-variance assumption $\sigma=0$ for the stochastic gradient (i.e., the deterministic gradient setting).
The improvement in convergence rates primarily results from our tighter bound on $\wsq(\nu,\nu^\lambda).$
We also note that both~\cite{brosse2017sampling,gurbuzbalaban2022penalized} focus on Euclidean projection, which is a special case within our framework. 

Furthermore, comparing the convergence rate of CLMC with that of CRLMC in Theorem~\ref{thm:rlmc}, we observe that the midpoint randomization yields a clear acceleration over the Euler discretization.

\subsection{Improved Error Bounds for Constrained Kinetic Langevin Monte Carlo}
\label{sec:klmc}

In this section, we revisit Langevin Monte Carlo algorithms in the constrained setting, focusing on the CKLMC method introduced in~\cite{gurbuzbalaban2022penalized}. This algorithm corresponds to a discretization of the SDE in~\eqref{KLD:2}, where the drift term $\nabla U^\lambda(\bL_t)$ is held fixed over each interval $[kh, (k+1)h)$ and replaced by $\nabla U^\lambda(\bL_{kh})$. More explicitly, the iterates of CKLMC are given by
\begin{align*}
\bv_{n+1}^{\sf CKLMC} & = (1 - \alpha \eta) \bv_n^{\sf CKLMC}
- \alpha \eta \nabla U^\lambda(\bvartheta_n^{\sf CKLMC})
+ \sqrt{2\gamma\eta} \sigma \bxi_n\\
\bvartheta_{n+1}^{\sf CKLMC} & = \bvartheta_n
+ \gamma^{-1} \eta\big(\alpha \bv_n^{\sf CKLMC}
- \beta \eta \nabla U^\lambda(\bvartheta_n^{\sf CKLMC})
+ \sqrt{2\gamma\eta} \tilde\sigma \bar\bxi_n\big),
\end{align*}
where
\begin{align*}
\alpha &= \frac{1 - e^{-\eta}}{\eta},
&\beta &= \frac{e^{-\eta} - 1 + \eta}{\eta^2} \\
\sigma^2 &= \frac{1 - e^{-2\eta}}{2\eta} ,
&\tilde{\sigma}^2 &= \frac{2(1 - 2\eta + 2\eta^2 - e^{-2\eta})}{(2\eta)^3},
\end{align*}
with $\eta=\gamma h$ and where $\bxi_n, \bar\bxi_n$ are independent $\mathcal N_p(\mathbf 0, \mathbf I_p)$ random vectors, independent of $(\bvartheta_n^{\sf CKLMC}, \bv_n^{\sf CKLMC})$.

It was shown in~\cite{gurbuzbalaban2022penalized} that this algorithm can effectively approximate the target distribution $\nu$. 
In the theorem below, we demonstrate that the refined analysis of $\wsq(\nu, \nu^\lambda)$ from Section~\ref{sec:general} yields improved upper bounds on the sampling error of CKLMC.
{
\begin{theorem}\label{thm:klmc}
Under Assumptions~\ref{asm:radius}-~\ref{asm:f1}, choose $\lambda \in \Bigl(0,\; \tfrac{\sqrt{c_1}\,r}{p+q} \wedge \tfrac{\sqrt{c_1}\,r\,e^{\operatorname{osc}_{\K}(f)}}{3\sqrt{\pi}\,\operatorname{Vol}(\K)\,p}\Bigr)$. 
Choose $\gamma$ and $h$ so that $\gamma\geqslant 5M^\lambda$ and $\sqrt{\kappa}\gamma h\leqslant 0.1$, where $\kappa=M^\lambda/m$.
Assume that $\bvartheta_0^{\sf CKLMC}$ is the minimizer of $U^\lambda$ and $\bv_0\sim\mathcal N(0,\gamma \bfI_p)$.
Then, for any $p>2$ and every $n\geqslant 1$, the distribution $\nu^{\sf CKLMC}_n$ of $\bvartheta^{\sf CKLMC}_n$ satisfies
\begin{align*}
    \wstwo(\nu_n^{\sf CKLMC},\nu)&\leqslant 2e^{-mhn}\wstwo(\nu_0^{\sf CKLMC},\nu)+0.9\gamma h\sqrt{\kappa p/m}+3C(p,2)\lambda^{\frac{1}{2}+\frac{1}{p}}\,,
\end{align*}
Furthermore, it holds that
\begin{align*}
    \wsone(\nu_n^{\sf CKLMC},\nu)\leqslant 2e^{-mhn}\sqrt{p/m}+0.9\gamma h\sqrt{\kappa p/m}+C(p,1)\lambda\,.
\end{align*}
\end{theorem}
}
Before discussing related work, we state the following corollary of the preceding theorem, obtained by taking $M^\lambda = M + M_0/\lambda^2$.
\begin{corollary}\label{cor:klmc}
Let the error level $\varepsilon\in(0,1)$ be small. Set $\gamma=5(M+M_0/\lambda^2).$ \\
noindent\textbf{(a)} Set $\lambda=\Theta(h^{2p/(7p+2)})$, and choose $h>0$ and $n\in\mathbb N_+$ so that
\begin{align*}
    h=\mathcal O\left(\varepsilon^{\frac{7p+2}{p+2}}\right)\quad\text{and}\quad n=\widetilde{\Omega}\left(\varepsilon^{-\frac{7p+2}{p+2}}\right)\,,
\end{align*}
then we have $\wstwo(\nu_n^{\sf CKLMC},\nu)=\logo(\varepsilon).$\\
\noindent\textbf{(b)} Set $\lambda=\Theta(h^{1/4})$, and choose $h>0$ and $n\in\mathbb N_+$ so that
\begin{align*}
    h=\mathcal O\left(\varepsilon^4\right)\quad\text{and}\quad n=\widetilde{\Omega}\left(\varepsilon^{-4}\right)\,,
\end{align*}
then we have $\wsone(\nu_n^{\sf CKLMC},\nu)=\logo(\varepsilon).$
\end{corollary}
When $p\geqslant 2$, we have $\frac{7p + 2}{p + 2} \geqslant 7 - \frac{6}{p}$. This allows us to simplify the requirement on the number of iterations $n$ needed to achieve $\wstwo(\nu_n^{\sf CKLMC}, \nu) = \logo(\varepsilon)$ to $n=\widetilde{\Omega}(\varepsilon^{-7+6/p}).$ 
A comparable convergence rate $\widetilde{\bigo}(\varepsilon^{-13})$ in $\wstwo$ distance can be derived from Proposition 2.22 in~\cite{gurbuzbalaban2022penalized} by setting the stochastic gradient variance $\sigma = 0$, which recovers the deterministic gradient setting considered in our work.
This improvement is primarily attributed to the tighter bound on $\wstwo(\nu,\nu^\lambda).$

A comparison of the convergence rates of CKLMC and CRKLMC in Theorem~\ref{thm:rklmc} reveals that the midpoint randomization scheme provides a clear acceleration over the Euler discretization.

\section{Relation to~\cite{brosse2017sampling} and~\cite{gurbuzbalaban2022penalized} }
\label{sec:comp}

Both works adopt a two-stage strategy similar to our work: they introduce a regularization term to construct a smooth surrogate density, which is then sampled using Langevin Monte Carlo (LMC) and/or kinetic Langevin Monte Carlo (KLMC). Below, we review their results in detail and highlight the key differences from our approach.

\subsection{Comparison with~\cite{brosse2017sampling}}

\cite{brosse2017sampling} investigates the same constrained sampling setting as in our work. They use the Moreau envelope (Example~\ref{ex:Maha} with $Q=I$) to construct a smooth surrogate density and then apply LMC to sample from it. Their results align naturally with our general framework, which combines Euclidean projection with LMC. However, their analysis focuses primarily on $\wsone$ and total variation (TV) distances.

In contrast, our work develops results for general $\wsq$ distances, accommodates a broader class of projection operators beyond the Euclidean case, and applies to a wider family of sampling algorithms beyond vanilla LMC. Moreover, our proof technique for Proposition~\ref{prop:w2_tight} can also recover the TV bound established in~\cite{brosse2017sampling}, achieving the same order. 
When combined with the standard TV convergence guarantees for LMC, this allows us to recover the TV convergence of CLMC as well.

\subsection{Comparison with~\cite{gurbuzbalaban2022penalized}}
After completing the first version of this work, we became aware of an updated version of~\cite{gurbuzbalaban2022penalized}, in which the authors analyze the behavior of LMC and KLMC for constrained sampling. Similar to~\cite{brosse2017sampling}, their approach relies on the Euclidean projection. However, their focus differs from ours in two key respects:

\begin{itemize}
    \item They analyze convergence in TV distance and allow the objective function $f$ to be smooth and possibly non-convex.
    \item They consider settings with access to stochastic gradients and derive $\wstwo$ error bounds for constrained LMC and KLMC. Specifically, when $f$ is strongly convex and smooth, the $\wass_2$ convergence rates are of order $\bigo(\varepsilon^{-18})$ for constrained LMC and $\bigo(\varepsilon^{-39})$ for constrained KLMC. 
They also extend their analysis of $\wstwo$ error bounds to the non-convex case, under smoothness assumptions on the potential $f$.
\end{itemize}
To obtain a comparable counterpart in their work, we consider the second setting where $f$ is strongly convex and smooth, and additionally assume that the stochastic gradient has zero variance, corresponding to the deterministic gradient setting considered in our analysis.
The corresponding convergence rates are detailed in Section~\ref{sec:lmc} and Section~\ref{sec:klmc}.
Our results significantly improve upon the convergence rates established in~\cite{gurbuzbalaban2022penalized}, with the improvement primarily due to a tighter bound on the Wasserstein distance between the surrogate density and the target distribution.

\section{Numerical Experiments}

In this section, we compare the performance of the CLMC, CKLMC, CRLMC, and CRKLMC algorithms using both synthetic and real data. 
The synthetic experiments, in particular, provide empirical support for our theoretical findings.
\subsection{Synthetic Experiment: 2D Scatter Plots}
Due to the difficulty of computing high-dimensional Wasserstein distances, we focus on a two-dimensional setting in this section to facilitate clearer presentation and easier visualization. Specifically, we consider the 2-dimensional standard normal distribution $\mathcal{N}(0, \bfI_2)$ and construct the target distribution $\nu$ by imposing two different choices for the constraint set $\K$.

The first setting uses a Euclidean ball centered at the origin, defined as $\K := \ball_2(0, r)$ with $r = 0.5$. In this case, we employ Euclidean projection, which coincides with the Gauge projection due to the symmetry of the constraint set.
The second setting significantly differs from the first. Here, $\K$ is defined as a 3-simplex shifted away from the origin 
\[
\K:=\{(x_1,x_2)\in\R^2:x_1,x_2\geqslant -0.3,x_1+x_2\leqslant 0.6\}.
\]
Here, we apply the Gauge projection to enforce the constraint.

\begin{figure}[h]
\centering
\includegraphics[width=1\linewidth]{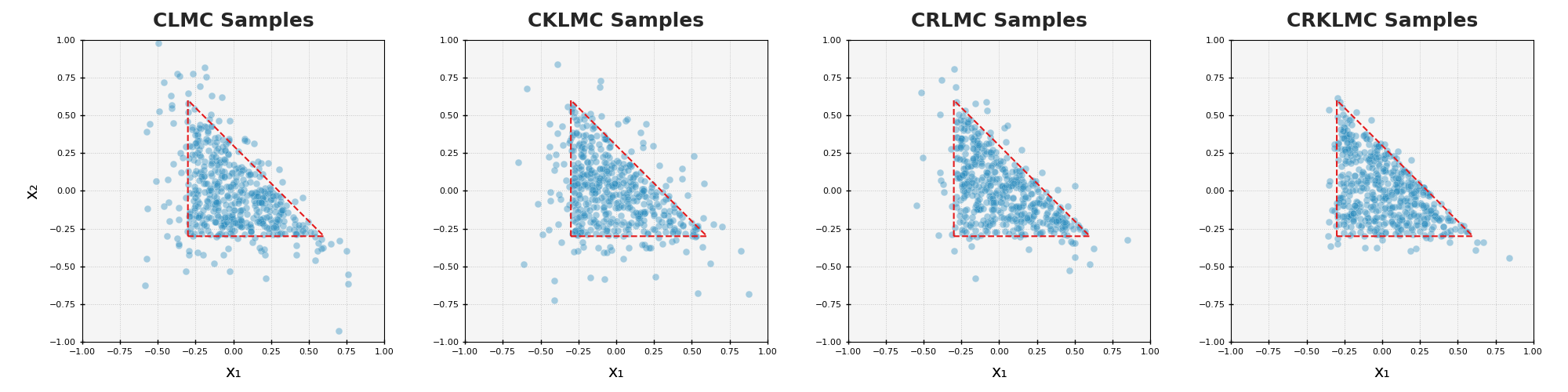}
\caption{Scatter plots of samples generated by $\{\text{CL,CKL,CRL,CRKL}\}\text{MC}$ algorithms.}
\label{fig:tria_less}
\end{figure}

\begin{figure}[h]
\centering
\includegraphics[width=1\linewidth]{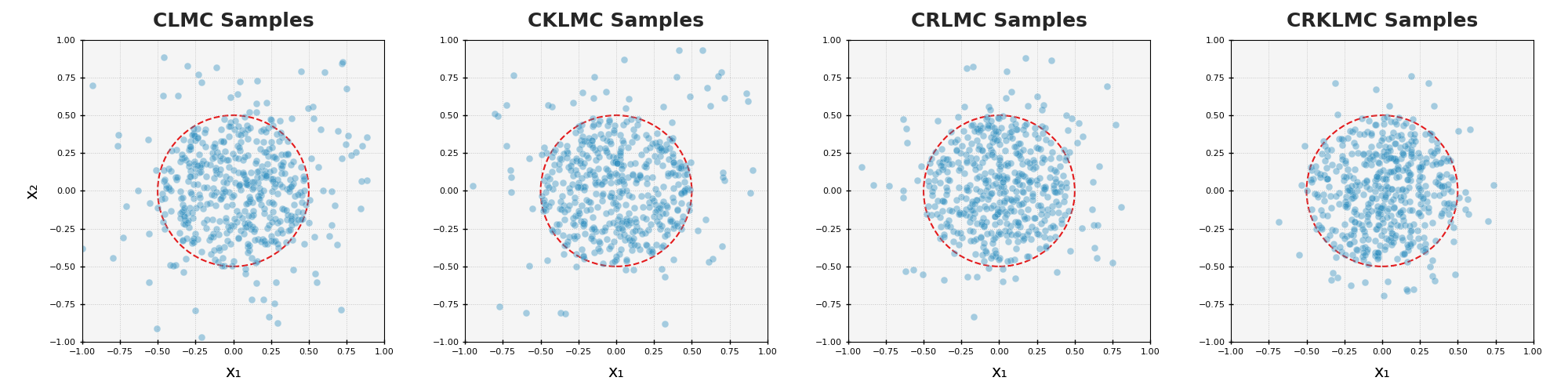}
\caption{Scatter plots of samples generated by $\{\text{CL,CKL,CRL,CRKL}\}\text{MC}$ algorithms.}
\label{fig:radial_less}
\end{figure}

Each algorithm is run for $n$ iterations. At the final iteration, $N$ samples are drawn from the approximate distribution to compare against the true target $\nu$. 
We use a step size of $h = 0.001$, and set the tuning parameters as follows
\[
\lambda^{\sf CLMC}=h^{1/4},~~\lambda^{\sf CRLMC}=h^{1/4},~~\lambda^{\sf CKLMC}=h^{3/10},~~\lambda^{\sf CRKLMC}=h^{3/8}\,.
\]
Figures~\ref{fig:tria_less} and~\ref{fig:radial_less} display scatter plots of the generated samples for both constraint sets with $n = 1000$ and $N = 500$. 
The red dashed lines in the plots indicate the boundaries of the constraint sets.
The visualizations offer valuable intuition and a qualitative sense of convergence. 
They indicate that CRKLMC achieves the best performance, with samples more tightly concentrated within the constraint set. CRLMC outperforms CLMC, and CRKLMC improves upon CKLMC, highlighting the advantage of midpoint randomization over Euler discretization. 
These observations align well with our theoretical findings.
Additional results with different combinations of $n$ and $N$, along with implementation details, are provided in Appendix~\ref{app:numerical}. 
The supplementary plots exhibit trends consistent with the observations discussed in this section

\subsection{Synthetic Experiments: Bayesian Lasso Regression}
\begin{figure}[h]
\centering
\begin{subfigure}[b]{1\linewidth}
    \centering
    \includegraphics[width=1\linewidth]{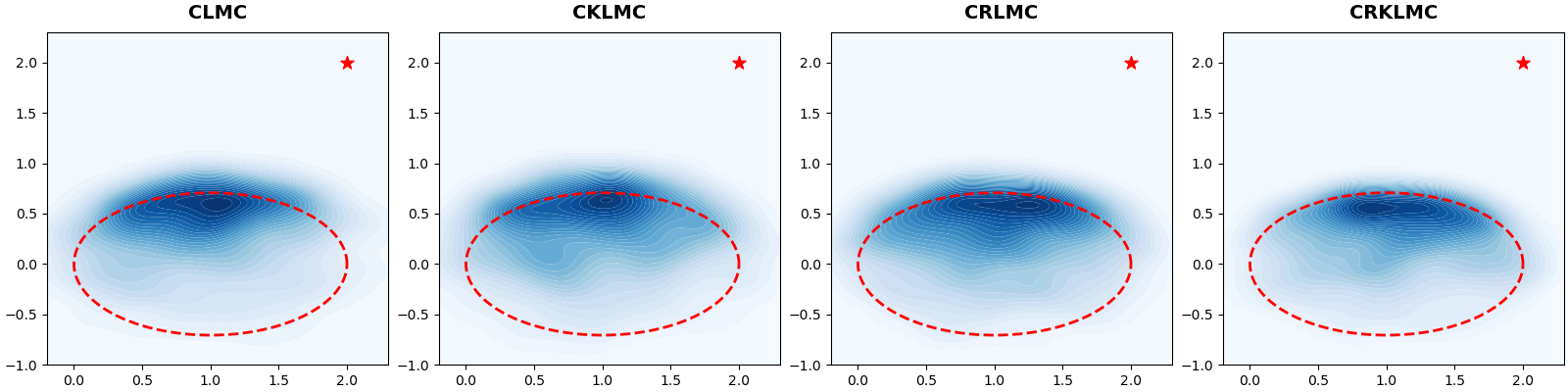}
    \caption{Contours generated by $\{\text{CL,CKL,CRL,CRKL}\}\text{MC}$ algorithms after 500 steps.}
    \label{fig:Bayes_case2_2000}
\end{subfigure}

\vspace{0.2cm}

\begin{subfigure}[b]{1\linewidth}
    \centering
    \includegraphics[width=1\linewidth]{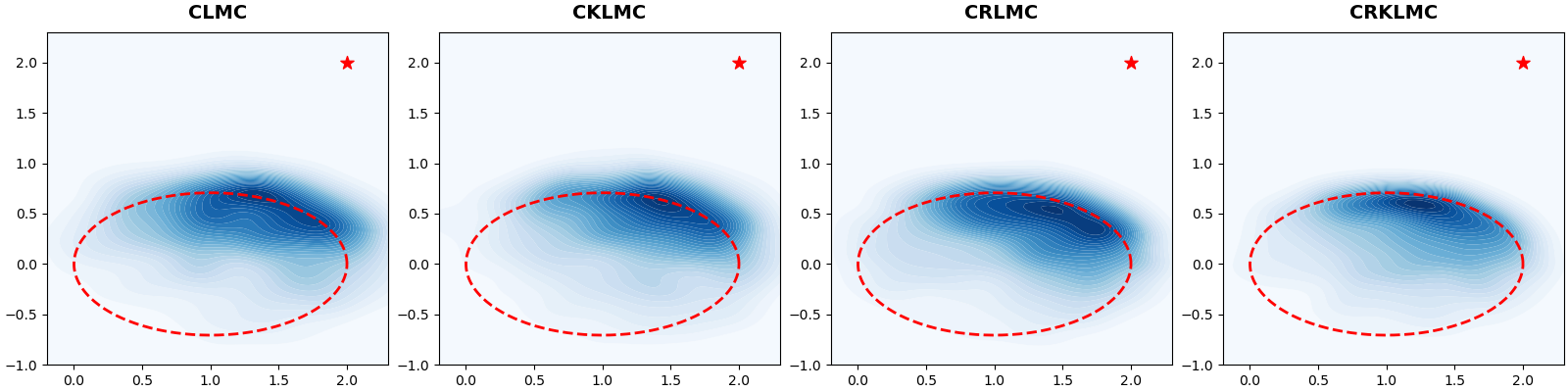}
    \caption{Contours generated by $\{\text{CL,CKL,CRL,CRKL}\}\text{MC}$ algorithms after 4000 steps.}
    \label{fig:Bayes_case2_4000}
\end{subfigure}

\caption{Contour plots generated by $\{\text{CL,CKL,CRL,CRKL}\}\text{MC}$ algorithms with different step sizes. The top panel shows contours after 2000 steps, while the bottom panel shows results after 4000 steps. The red star denotes the true solution.}
\label{fig:Bayes_case2_combined}
\end{figure}

We next consider a Bayesian Lasso regression model.
We generate $10000$ data points $(a_j,y_j)$ according to 
\begin{align*}
    \delta_j\sim\mathcal N(0,0.25),\quad a_j\sim\mathcal N(0,\mathbf{I}_2),\quad y_j=x_*^\top a_j+\delta_j,\quad x_*=[2,2]^\top.
\end{align*}
The constraint set is taken to be the ellipsoid
\begin{align*}
    \mathcal K:=\{x:(x-\bar a_1)^\top Q_1(x-\bar a_1)\leqslant \bar b_1\},
\end{align*}
where we set $\bar a_1=[1,0]^\top,\bar b_1=1$, and 
\begin{align*}
    Q_1=\begin{pmatrix}1&0\\0&2\end{pmatrix}
\end{align*}
We adopt a uniform prior supported on $\K$
The posterior distribution is therefore
\begin{align*}
    \nu(x)\propto e^{\sum_{j=1}^n-\frac{1}{2}(y_j-x^\top a_j)^2}\cdot\mathbbm 1_\mathcal K,
\end{align*}
where $\mathbbm 1_\mathcal K$ is the indicator function for the constraint set $\mathcal K$ and $n$ is the number of observations. 
Each algorithm is run for $n=2000,4000$ iterations, generating $N=1000$ samples, with initialization drawn from the standard Gaussian distribution.
The tuning parameters and other settings match those used in the earlier experiments. Figure~\ref{fig:Bayes_case2_combined} shows contour plots constructed from the sampled points. 
Comparing the first and second rows reveals the differing convergence speeds of the algorithms toward the target posterior distribution.
{Across these contour plots, CRKLMC converges the fastest: its distribution at 500 steps already closely matches the result at 4000 step . In contrast, CLMC converges the slowest, showing the largest shape change between 500 and 4000 steps, while CKLMC and CRLMC lie in between with more moderate changes.}

\subsection{Real Data: UCI Adult dataset}
In this section, we apply the four constrained sampling algorithms to the UCI Adult dataset~\cite{adult_2}. Predictive models in modern machine learning often inherit societal biases present in training data, leading to unequal outcomes across demographic groups. A central challenge is to balance predictive performance with algorithmic fairness. By imposing fairness constraints directly on the model parameters, our goal is to obtain posterior distributions that satisfy fairness requirements while retaining reasonable predictive power, providing a Bayesian perspective on fair inference.

The UCI Adult dataset includes demographic and socioeconomic features. 
Our setup includes:\\
$\bullet$~~\textbf{Sensitive attribute}: \texttt{sex} (male or female), used to define the protected groups.\\
$\bullet$~~\textbf{Non-sensitive features}: numerical covariates including \texttt{age}, \texttt{education-num}, \texttt{fnlwgt}, \texttt{capital-gain}, \texttt{capital-loss}, and \texttt{hours-per-week}, standardized for consistent scaling.\\
$\bullet$~~\textbf{Target variable}: a binary income indicator (\texttt{>50k} as $1$; \texttt{$\leqslant$ 50k} as 
$0$).

We first compute the unconstrained maximum likelihood estimate (MLE) $\btheta_{\text{uncon}}$ of the logistic regression, which provides a baseline achieving high predictive accuracy but does not enforce any fairness constraints. This unconstrained estimate also determines the scale of the fairness and norm constraints described below.

To impose fairness, we introduce two convex constraints. The first limits disparities in predicted outcomes between males and females
\begin{align*}
    \K_1=\{\btheta\in\R^d:|(\mu_{\text{male}}-\mu_{\text{female}})^\top\btheta|<\alpha\Delta_0\}
\end{align*}
where $\mu_{\text{male}}$ and $\mu_{\text{female}}$ denote the sample mean of non-sensitive features. $\btheta$ is the parameter subvector associated with those features. $\Delta_0=|(\mu_{\text{male}}-\mu_{\text{female}})^\top\btheta_{\text{uncon}}|$ measures the baseline disparity under the unconstrained MLE. We set $\alpha=0.5$, which adjusts the strictness of the fairness constraint. 

The second constraint,
\begin{align*}
    \K_2=\{\btheta\in\R^d:\|\btheta\|_2<3\|\btheta_{\text{uncon}}\|_2\}.
\end{align*}
prevents overfitting and stabilizes the posterior by restricting the parameter norm. We define the overall feasible set as $\K=\K_1\cap \K_2.$

Similar fairness constraints have been studied in recent work on constrained sampling~\cite{chamon2024constrained}, which employs a Primal–Dual Langevin Monte Carlo approach.
In contrast, our formulation imposes fairness directly through a convex support constraint, which enforces individual-level parity. This differs fundamentally from their approach, which ensures fairness only in expectation across groups rather than at the level of each sampled parameter vector.

\begin{figure}
    \centering
    \includegraphics[width=0.65\linewidth]{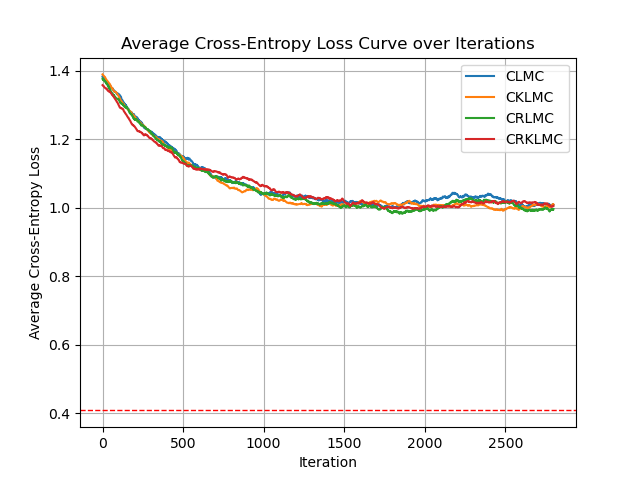}
    \caption{Average Cross-Entropy Loss Convergence of Constrained LMC Variants Over Iterations}
    \label{fig:adult}
\end{figure}
To evaluate convergence, we track the average cross-entropy loss along the sampling trajectories. As shown in Figure~\ref{fig:adult}, all four algorithms display similar behavior: the loss decreases rapidly during early iterations and eventually stabilizes. This pattern indicates that each algorithm successfully samples from a well-defined posterior distribution over $\btheta$ consistent with the fairness and norm constraints. The stability of the loss curves confirms that the resulting samples accurately reflect the intended balance between fairness and predictive performance.

\section{Discussion}

In this work, we use the Wasserstein distance as the primary metric for quantifying sampling error, owing to its natural connection with optimal transport theory. Recent developments in gradient-based sampling, however, have considered alternative metrics such as TV distance, KL divergence, and $\chi^2$ divergence. Our proof technique for bounding $\wsq(\nu, \nu^\lambda)$ can in fact be extended to obtain TV bounds, which would allow us to characterize the behavior of constrained sampling algorithms once a TV guarantee is available for both the algorithm’s output and the surrogate distribution $\nu^\lambda$. 
Extending existing convergence guarantees for constrained sampling to metrics like KL divergence and $\chi^2$ divergence is a promising direction for future work.

Moreover, although our current analysis focuses on Langevin-type methods in constrained settings with full gradient information, our theoretical framework is general and can be seamlessly adapted to other sampling algorithms, provided that the error bound of the sampling error has explicit parameter dependence.

Finally, we note that strong convexity and smoothness of the potential $f$ are required only for analyzing the four Langevin-type sampling algorithms. 
In contrast, the assumptions used to control the approximation error between $\nu^\lambda$ and $\nu$ are much weaker and do not rely on the smoothness and convexity of $f$. 
Therefore, once the convexity requirement in the sampling error analysis can be relaxed, our framework would enable the study of constrained sampling algorithms in fully nonconvex settings as well.

\section{Acknowledgments}
We are grateful to Arnak Dalalyan for inspiring discussions and constructive comments during the early stages of this manuscript. Part of this work was carried out while Lu Yu was a postdoctoral researcher at CREST/ENSAE Paris. Lu Yu is supported by the City University of Hong Kong Startup Fund and the Hong Kong RGC ECS Grant 21306325.

\bibliographystyle{amsalpha}
\bibliography{bib}

\appendix

\section{Proofs of Section~\ref{sec:approx_target}}
In this part, we provide the proof of Proposition~\ref{prop:w2_tight} and Proposition~\ref{prop:lb}.\\

\noindent \textbf{Additional Notation.} Throughout the remainder, 
$\mathbb{S}_{p-1}$ denotes the unit sphere in $\R^p,$ 
$\omega_{p-1}$ denotes the $(p-1)$-dimensional surface area of $\mathbb{S}_{p-1}$, and $V_p$ the volume of the $p$-dimensional unit ball.

\subsection{Proof of Proposition~\ref{prop:w2_tight}}
\label{app:prop1}
Before presenting the formal proof, we outline the main idea.
The key ingredient is a transport map $T$ that allows us to decompose the Wasserstein distance as
\[
\wsq(\nu,\nu^\lambda)\leqslant \wsq(\nu,T_{\#}\nu^\lambda)+\wsq(T_{\#}\nu^\lambda,\nu^\lambda)\,,
\]
and analyze each term separately.  
The first term accounts for the discrepancy between $\nu$ and $T_{\#}\nu$, which only arises in a small neighborhood near the origin. 
Outside this region, the transport map $T$ exactly aligns the two distributions. 
The second term captures the cost of transporting $\nu^\lambda$ under the map $T$, and is decomposed into three radial regions: an interior region near the origin, 
a near-boundary region where the discrepancy in radial coordinates is small but non-negligible, and an exterior region outside the set $\K$.
This structured analysis leads to sharp, non-asymptotic bounds on the Wasserstein distance.

\begin{proof}[Proof of Proposition~\ref{prop:w2_tight}]
In polar coordinates, any vector $x\in\R^p$ can be uniquely expressed as 
$x=\rho\theta$, where $\rho\geqslant 0$ is the radial coordinate and $\theta\in\mathbb S_{p-1}$ denotes the directional unit vector on the $(p-1)$-dimensional sphere.
For each direction $\theta$, the radial function is defined as
\[
R(\theta):=\sup\{r\geqslant 0:r\theta\in\K\}\,.
\]
which characterizes the extent of the set $\K$ along direction $\theta.$
This parameterization allows us to define the radial cumulative distribution function accordingly.
\begin{align*}
    F(\rho,\theta)&:=\dfrac{\int_0^{\rho} t^{p-1}e^{-f(t\theta)}\rmd t}{\int_{\K}e^{-f(x)}\rmd x},\quad \rho\in[0,R(\theta)]\,,\\
    F_\lambda(\rho,\theta)&:=\dfrac{\int_0^\rho t^{p-1}e^{-f(t\theta)-\frac{1}{2\lambda^2}d_{\K}(t\theta)}\rmd t}{\int_{\K}e^{-f(x)}\rmd x+\int_{\K^c}e^{-f(x)-\frac{1}{2\lambda^2}d_{\K}(x)}\rmd x},\quad \rho\in[0,\infty)\,.
\end{align*}
Define
\begin{align*}
    Z:=\int_{\K}e^{-f(x)}\rmd x\quad \text{and}\quad Z_{\lambda}:=Z+\int_{\K^c}e^{-U^{\lambda}(x)}\rmd x\,.
\end{align*}
To bound the difference $Z_\lambda-Z$, we need the following lemma.

\begin{lemma}\label{lem:ext_int}
Under Assumptions~\ref{asm:radius}-\ref{asm:dK}, for any $k\geqslant 0$, if $\lambda<c_1^{1/2}r/(p+k)$, it holds that
    \begin{align*}
        \int_{\K^c}\normtwo{x}^ke^{-U^\lambda(x)}\rmd x\leqslant\dfrac{3\sqrt{\pi}(p+k)e^{-l}}{rc_1^{1/2}}\lambda\times \int_{\K}\normtwo{x}^k\rmd x\,.
    \end{align*}
\end{lemma}
By Lemma~\ref{lem:ext_int}, we have
\begin{align}
    Z_{\lambda}-Z=\int_{\K^c}e^{-U^\lambda(x)}\rmd x\leqslant \dfrac{3\sqrt{\pi}pe^{-l}\text{Vol}(\K)}{rc_1^{1/2}}\lambda\,.
    \label{eq:Zdiff}
\end{align}
Define $\epsilon(\theta):=F_\lambda(\infty,\theta)-F(R(\theta),\theta).$ 
The following lemma provides a bound for this term.
\begin{lemma}
    \label{lem:epsilon}
    Assume the assumptions stated in Lemma~\ref{lem:ext_int} are satisfied with $k=0$. There exists a universal constant $C_\epsilon(p)>0$ such that for all $\theta\in\mathbb{S}_{p-1}$, it holds that $|\epsilon(\theta)|\leqslant C_\epsilon(p)\lambda,$ where $C_\epsilon(p)=\frac{5e^{2\omega(f,\K)}}{c_1^{1/2}}(a_1p)^{(p+1)/2}$, and $a_1=\frac{\max(R^2,1)}{r^2}$ is constant independent of $p$ and $\lambda.$ 
\end{lemma}
By Implicit Function Theorem, there exist two differentiable functions $g$ and $g_\lambda:[0,\infty)\times\mathbb S_{p-1}\to[0,\infty)$, satisfying
\begin{align*}
    F(g(c,\theta),\theta)=c,\quad\text{and}\quad F_\lambda(g_\lambda(c,\theta),\theta)=c
\end{align*}
respectively. 
\begin{lemma}\label{lem:gF}
Denote the upper bound of $f$ on $\K$ by $u$, that is $l\leqslant f(x)\leqslant u,\forall x\in\K$. 
The following three properties hold for the functions $F, F_\lambda$, and their inverse functions $g,g_\lambda$.
    \begin{enumerate}[leftmargin=10pt]
    \item For any fixed $\theta\in\mathbb S_{p-1},$ $g(x,\theta)$ and $g_\lambda(x,\theta)$ are both increasing functions with respect to $x$.
    \item For $\rho\leqslant R(\theta)$, we have 
    \[\rho^pe^{-u}/(pZ)\leqslant F(\rho,\theta)\leqslant \rho^pe^{-l}/(pZ)
    \]
    and 
    \[\rho^pe^{-u}/(pZ_\lambda)\leqslant F_\lambda(\rho,\theta)\leqslant \rho^pe^{-l}/(pZ_\lambda)\,.
    \] 
    Thus, it holds that
    \begin{align*}
        x^{1/p}(pZe^l)^{1/p}\leqslant g(x,\theta)\leqslant x^{1/p}(pZe^u)^{1/p}
    \end{align*}
    and
    \begin{align*}
        x^{1/p}(pZ_\lambda e^l)^{1/p}\leqslant g_\lambda(x,\theta)\leqslant x^{1/p}(pZ_\lambda e^u)^{1/p}\,.
    \end{align*}
    \item For $\rho\leqslant R(\theta)$, we have $F(\rho,\theta)=F_\lambda(\rho,\theta)\cdot Z_\lambda/Z$, therefore, it holds that
    \begin{align*}
        g(x,\theta)=g_\lambda(x\cdot Z/Z_\lambda,\theta),\quad \forall x\leqslant F(R(\theta),\theta)\,.
    \end{align*}
    \end{enumerate}
\end{lemma}
Define a deterministic transport map $T$ of the form
\begin{align*}
    T(\rho\theta)=g(\max\{F_\lambda(\rho,\theta)-\epsilon(\theta),0\},\theta)\theta\,.
\end{align*}
We decompose the Wasserstein distance between $\nu$ and $\nu^\lambda$ into two components
\begin{equation}
\wsq(\nu,\nu^\lambda)\leqslant\wsq(\nu,T_{\#}(\nu^\lambda))+\wsq(T_{\#}(\nu^\lambda),\nu^\lambda)\,.
\label{eq:decomp}
\end{equation}
In the following, we derive upper bounds for the two terms on the right-hand side.

Define two sets centered around the origin as 
\[
\K_0:=\{\rho\theta:\rho\leqslant g(\max\{-\epsilon(\theta),0\},\theta)\}
\]
and 
\[\K_{0,\lambda}:=\{\rho\theta:\rho\leqslant g_\lambda(\max\{0,\epsilon(\theta)\},\theta)\}\,.
\]
We note that
\[
T_{\#}(\nu^\lambda\big|_{\K_{0,\lambda}^c})=\nu\big|_{\K_0^c}.
\]
The first term on the right-hand side of display~\eqref{eq:decomp} can then be bounded by
\begin{equation}
\label{eq:term1}
    \wsq^q(\nu,T_{\#}(\nu^\lambda))\leqslant\int_{\K_0}\text{diam}(\K_0)^q\nu(\rmd x)=\text{diam}(\K_0)^q\nu(\K_0)
\end{equation}
We note that 
\[
\text{diam}(\K_0)\leqslant 2\sup_{\theta\in\mathbb S_{p-1}}g(\max\{-\epsilon(\theta),0\},\theta)\leqslant 2\sup_{\theta\in\mathbb S_{p-1}}g(|\epsilon(\theta)|,\theta).
\]
By Lemma~\ref{lem:epsilon} and Lemma~\ref{lem:gF}, we have
\begin{align*}
    g(|\epsilon(\theta)|,\theta)\leqslant |\epsilon(\theta)|^{1/p}(pZe^u)^{1/p}\leqslant \lambda^{1/p}(pC_\epsilon(p)Ze^u)^{1/p}\,.
\end{align*}
This implies
\[
\text{diam}(\K_0)\leqslant 2\lambda^{1/p}(pC_\epsilon(p)Ze^u)^{1/p}\,.
\]
Moreover, it holds that 
\[\nu(\K_0)\leqslant e^{-l}\text{Vol}(\K_0)/Z\leqslant e^{-l}V_p\,\text{diam}(\K_0)^p/Z\leqslant e^{\omega(f,\K)}\text{diam}(\K_0)^p/r^p,
\]
where $\omega(f,\K):=\sup_{x\in\K}f(x)-\inf_{x\in\K}f(x)$ denotes the oscillation of $f$ over $\K$. 
Plugging these back to the previous display~\eqref{eq:term1} yields
\begin{align*}
    \wsq^q(\nu,T_{\#}(\nu^\lambda))&\leqslant e^{\omega(f,\K)}/r^p\cdot\text{diam}(\K_0)^{p+q}\\
    &\leqslant \lambda^{(p+q)/p} e^{\omega(f,\K)}/r^p\cdot 2^{p+q}(pC_\epsilon(p)Ze^u)^{(p+q)/p}\,.
\end{align*}
Since $Z\leqslant e^{-l}R^pV_p\leqslant e^{-l+1/6}R^p(2\pi e)^{p/2}/(\sqrt{\pi}p^{(p+1)/2}),$ we finally have 
\begin{align*}
    \wsq^q(\nu,T_{\#}(\nu^\lambda))&\leqslant e^{\omega(f,\K)}\lambda^{(p+q)/p}\dfrac{2^{p+q}(5e^{2\omega(f,\K)}a_1^{(p+1)/2}pe^{\omega(f,\K)+1/6}R^p(2\pi e)^{p/2})^{(p+q)/p}}{r^p(c_1^{1/2}\sqrt{\pi})^{(p+q)/p}}\\
    &\leqslant p^{(p+q)/p}\lambda^{(p+q)/p}\cdot\left(\dfrac{2\cdot 5^{1/p}\sqrt{2\pi e}e^{4\omega(f,\K)/p+1/(6p)}a_1^{(p+1)/(2p)}R}{\min(r,1)c_1^{1/(2p)}\sqrt{\pi}^{1/p}}\right)^{p+q}\\
    &\leqslant a_2^{p+q}p^{(p+q)/p}\lambda^{(p+q)/p}\,,
\end{align*}
where $a_2=\frac{28 e^{4\omega(f,\K)/p}}{c_1^{1/(2p)}}\cdot\left(\frac{\max(R,1)}{\min(r,1)}\right)^{(2p+1)/p}.$ 

We now aim to establish the upper bound for~$\wsq(T_{\#}(\nu^\lambda),\nu^\lambda)$.
Note that 
\begin{align*}
    \wsq^q(T_{\#}(\nu^\lambda),\nu^\lambda)&\leqslant\int_{\mathbb S_{p-1}}\int_0^{\infty}\normtwo{t\theta-T(t\theta)}^q\dfrac{t^{p-1}e^{-f(t\theta)-\frac{1}{2\lambda^2}d_{\K}(t\theta)}}{\int_{\K}e^{-f(x)}\rmd x+\int_{\K^c}e^{-f(x)-\frac{1}{2\lambda^2}d_{\K}(x)}\rmd x}\rmd t\,\rmd\theta\\
    &=\int_{\mathbb S_{p-1}}\underbrace{\int_0^\infty|t-g(\max\{F_{\lambda}(t,\theta)-\epsilon(\theta),0\},\theta)|^q\dfrac{\partial F_\lambda(t,\theta)}{\partial t}\rmd t}_{\phi(\theta)}\rmd \theta\,.
\end{align*}
For any fixed $\theta\in\mathbb S_{p-1}$, it holds that
\begin{align*}
    \phi(\theta)&=\underbrace{\int_0^{g_\lambda(\max\{0,\epsilon(\theta)\},\theta)}|t-g(0,\theta)|^q\dfrac{\partial F_\lambda(t,\theta)}{\partial t}\rmd t}_{\phi_1(\theta)}\\
    &+\underbrace{\int_{g_\lambda(\max\{0,\epsilon(\theta)\},\theta)}^{R(\theta)}|t-g(F_\lambda(t,\theta)-\epsilon(\theta),\theta)|^q\dfrac{\partial F_\lambda(t,\theta)}{\partial t}\rmd t}_{\phi_2(\theta)}\\
    &+\underbrace{\int_{R(\theta)}^\infty |t-g(F_\lambda(t,\theta)-\epsilon(\theta),\theta)|^q\dfrac{\partial F_\lambda(t,\theta)}{\partial t}\rmd t}_{\phi_3(\theta)}.
\end{align*}
In the following, we derive upper bounds for the three terms $\phi_1,\phi_2$ and $\phi_3$.

\noindent\textbf{Step 1.} Since $g(0,\theta)=0$, we have that
\begin{align*}
    \int_{\mathbb S_{p-1}}\phi_1(\theta)\rmd \theta&=\int_{\mathbb S_{p-1}}\int_0^{g_\lambda(\max\{0,\epsilon(\theta)\},\theta)}t^q\dfrac{t^{p-1}e^{-f(t\theta)-\frac{1}{2\lambda^2}d_{\K}(t\theta)}}{\int_{\K}e^{-f(x)}\rmd x+\int_{\K^c}e^{-f(x)-\frac{1}{2\lambda^2}d_{\K}(x)}\rmd x}\rmd t\,\rmd\theta\\
    &=\int_{\K_{0,\lambda}}\normtwo{x}^q\nu^\lambda(\rmd x)\\
    &\leqslant \text{diam}(\K_{0,\lambda})^q\nu^\lambda(\K_{0,\lambda})\\
    &\leqslant \text{diam}(\K_{0,\lambda})^qe^{-l}\text{Vol}(\K_{0,\lambda})/Z_\lambda\\
    &\leqslant \text{diam}(\K_{0,\lambda})^{p+q}e^{\omega(f,\K)}/r^p\,.
\end{align*}
Note that
\[
\text{diam}(\K_{0,\lambda})\leqslant 2\sup_{\theta\in\mathbb S_{p-1}}g_\lambda(|\epsilon(\theta)|,\theta)\leqslant 2\lambda^{1/p}(pC_\epsilon(p)Z_\lambda e^u)^{1/p} \,.
\]
In addition, when $\lambda<\frac{e^{-\omega(f,\K)}}{3\sqrt{\pi}\text{Vol}(\K)}\cdot\frac{rc_1^{1/2}}{p}$, it holds that $Z_\lambda<2Z.$ Therefore, we obtain
\begin{align*}
    \int_{\mathbb S_{p-1}}\phi_1(\theta)\rmd \theta&\leqslant 2^{p+q}(2pC_\epsilon(p)Ze^u)^{(p+q)/p}e^{\omega(f,\K)}/r^p\times\lambda^{(p+q)/p}\\
      &\leqslant (2^{1/p}a_2)^{p+q}p^{(p+q)/p}\lambda^{(p+q)/p}\,.
\end{align*}
\noindent\textbf{Step 2.} We now establish the upper bound for $\phi_2(t)$. 
Let $y=F_\lambda(t,\theta)$. For ease of the notation, we denote the function $h(\cdot,\theta)$ with fixed $\theta$ by $h(\cdot|\theta)$. Note that $g(\cdot|\theta)=F^{-1}(\cdot|\theta)$ and $g_\lambda(\cdot|\theta)=F_\lambda^{-1}(\cdot|\theta).$ 
Using the change of variable, we have 
\begin{align*}
    \phi_2(\theta)&=\int_{\max\{0,\epsilon(\theta)\}}^{F_\lambda(R(\theta)|\theta)}\left|F_\lambda^{-1}(y|\theta)-F^{-1}(y-\epsilon(\theta)|\theta)\right|^q\rmd y\\
    &=\int_{\max\{0,\epsilon(\theta)\}}^{F_\lambda(R(\theta)|\theta)}\left|F^{-1}(y\cdot Z_\lambda/Z|\theta)-F^{-1}(y-\epsilon(\theta)|\theta)\right|^q\rmd y\,.
\end{align*}
The last equality follows from Lemma~\ref{lem:gF}. 
Then, we decompose the integral into two parts
\begin{align*}
    \phi_2(\theta)&=\int_{\max\{0,\epsilon(\theta)\}}^{2C_\epsilon(p)\lambda}\left|F^{-1}(y\cdot Z_\lambda/Z|\theta)-F^{-1}(y-\epsilon(\theta)|\theta)\right|^q\rmd y\\
    &\quad +\int_{2C_\epsilon(p)\lambda}^{F_\lambda(R(\theta)|\theta)}\left|F^{-1}(y\cdot Z_\lambda/Z|\theta)-F^{-1}(y-\epsilon(\theta)|\theta)\right|^q\rmd y\,.
\end{align*}
By Lemma~\ref{lem:gF} and an application of the mean value theorem to $F^{-1}$, we obtain
\begin{align*}
\phi_2(\theta)
    &\leqslant\underbrace{\int_{\max\{0,\epsilon(\theta)\}}^{2C_\epsilon(p)\lambda}\left|(y\cdot Z_\lambda/Z)^{1/p}(pZe^u)^{1/p}+(y-\epsilon(\theta))^{1/p}(pZe^u)^{1/p}\right|^q\rmd y}_{I_1(\theta)}\\
    &\quad+\underbrace{\int_{2C_\epsilon(p)\lambda}^{F_\lambda(R(\theta)|\theta)}\left|y\cdot\dfrac{Z_\lambda-Z}{Z}+\epsilon(\theta)\right|^q\left|\dfrac{\rmd F^{-1}(z|\theta)}{\rmd z}\bigg|_{z=\xi}\right|^q\rmd y}_{I_2(\theta)}\,.
\end{align*}
We have
\begin{align*}
    I_1(\theta)&\leqslant 2C_\epsilon(p)\lambda\times |2(4C_\epsilon(p)\lambda)^{1/p}(pZe^u)^{1/p}|^q\\
    &\leqslant \lambda^{(p+q)/p}\times 2^{q+1}C_\epsilon(p)(4pC_\epsilon(p)Ze^u)^{q/p}\\
    &=\lambda^{(p+q)/p}\times 2^{q+1}\left[\dfrac{2e^{2\omega(f,\K)}}{c_1^{1/2}}(a_1p)^{(p+1)/2}\right]^{(p+q)/p}(4pZe^u)^{q/p}\\
    &\leqslant \lambda^{(p+q)/p}p^{(p+q)/p+(p-1)/2}\times 2^{q+1}\left[\dfrac{2e^{2\omega(f,\K)}a_1^{(p+1)/2}}{c_1^{1/2}}\right]^{\frac{p+q}{p}}\left[\dfrac{4e^{\omega(f,\K)+1/6}R^p(2\pi e)^{p/2}}{\sqrt{\pi}}\right]^{\frac{q}{p}}\,.
   \end{align*}
Therefore
\begin{align*}
    &\quad\int_{\mathbb S_{p-1}}I_1(\theta)\rmd \theta\\
    &\leqslant \dfrac{2\pi^{p/2}}{\Gamma(p/2)}\sup_{\theta\in\mathbb S_{p-1}}I_1(\theta)\\
    &\leqslant \lambda^{(p+q)/p}p^{(p+q)/p}\times\dfrac{(2\pi e)^{p/2}}{\sqrt{\pi}}\times 2^{q+1}\left[\dfrac{2e^{2\omega(f,\K)}a_1^{(p+1)/2}}{c_1^{1/2}}\right]^{\frac{p+q}{p}}\left[\dfrac{4e^{\omega(f,\K)+1/6}R^p(2\pi e)^{p/2}}{\sqrt{\pi}}\right]^{\frac{q}{p}}\\
    &\leqslant \lambda^{(p+q)/p}p^{(p+q)/p}\times\left[\dfrac{2^{1+3/p}\sqrt{2\pi e}e^{3\omega(f,\K)/p}}{c_1^{1/(2p)}}\cdot\left(\dfrac{\max(R,1)}{\min(r,1)}\right)^{(p+1)/p+q/(p+q)}\right]^{p+q}\\
    &\leqslant (2^{2/p}a_2)^{p+q}p^{(p+q)/p}\lambda^{(p+q)/p}\,.
\end{align*}
Note that $F_\lambda(R(\theta)|\theta)<F(R(\theta)|\theta)\leqslant R^pe^{-l}/(pZ)$ and we also have 
\eqn{
Z\geqslant e^{-u}r^pV_p\geqslant e^{-u}r^p(2\pi e)^{p/2}/(\sqrt{2\pi}p^{(p+1)/2}),} 
then it holds that,
\begin{align*}
    I_2(\theta)&\leqslant \lambda^q\left|F_\lambda(R(\theta)|\theta)\cdot\dfrac{3\sqrt{\pi}pe^{-l}\text{Vol}(\K)}{rc_1^{1/2}Z}+C_\epsilon(p)\right|^q \int_{2C_\epsilon(p)\lambda}^{F_\lambda(R(\theta)|\theta)}\left|\dfrac{\rmd F^{-1}(z|\theta)}{\rmd z}\bigg|_{z=\xi}\right|^q\rmd y\\
    &\leqslant\lambda^q\left|\dfrac{3\sqrt{\pi}R^pe^{-2l}\text{Vol}(\K)}{rc_1^{1/2}Z^2}+C_\epsilon(p)\right|^q\int_{2C_\epsilon(p)\lambda}^{F_\lambda(R(\theta)|\theta)}\left|\dfrac{\rmd F^{-1}(z|\theta)}{\rmd z}\bigg|_{z=\xi}\right|^q\rmd y\\
    &\leqslant\lambda^q\left|\dfrac{3\sqrt{\pi}R^pe^{2\omega(f,\K)}}{r^{p+1}c_1^{1/2}V_p}+C_\epsilon(p)\right|^q\int_{2C_\epsilon(p)\lambda}^{F_\lambda(R(\theta)|\theta)}\left|\dfrac{\rmd F^{-1}(z|\theta)}{\rmd z}\bigg|_{z=\xi}\right|^q\rmd y\\
    &\leqslant\lambda^q\left|\dfrac{3\sqrt{2}\pi R^pe^{2\omega(f,\K)}}{r^{p+1}c_1^{1/2}(2\pi e)^{p/2}}p^{(p+1)/2}+C_\epsilon(p)\right|^q\int_{2C_\epsilon(p)\lambda}^{F_\lambda(R(\theta)|\theta)}\left|\dfrac{\rmd F^{-1}(z|\theta)}{\rmd z}\bigg|_{z=\xi}\right|^q\rmd y\\
    &\leqslant \lambda^q\cdot |2C_\epsilon(p)|^q\int_{2C_\epsilon(p)\lambda}^{F_\lambda(R(\theta)|\theta)}\left|\dfrac{\rmd F^{-1}(z|\theta)}{\rmd z}\bigg|_{z=\xi}\right|^q\rmd y
\end{align*}
where $\xi\in[\min\{y\cdot Z_\lambda/Z,y-\epsilon(\theta)\},\max\{y\cdot Z_\lambda/Z,y-\epsilon(\theta)\}]$. 
Taking the derivative of $F^{-1}$ gives 
\begin{align*}
    \dfrac{\rmd F^{-1}(z|\theta)}{\rmd z}=\dfrac{1}{F^\prime(F^{-1}(z|\theta)|\theta)}=\dfrac{Ze^{f(F^{-1}(z|\theta)\theta)}}{F^{-1}(z|\theta)^{p-1}}\,.
\end{align*}
Recall that $f(x) \leqslant u$ on $\K$.
Notice that $F^{-1}(z|\theta)\leqslant R(\theta)$ since 
\[
z\leqslant\max\{F_\lambda(R(\theta)|\theta)\cdot Z_\lambda/Z,F_\lambda(R(\theta)|\theta)-\epsilon(\theta)\}\leqslant F(R(\theta)|\theta),
\]
By the monotonicity of $F$, this implies $F^{-1}(z|\theta)\leqslant R(\theta)$.
Consequently, since $f(x) \leqslant u$ for all $x\in\K$, it follows that 
\[
e^{f(F^{-1}(z|\theta)\theta)}\leqslant e^u\,.
\]
Moreover, we have $\xi\geqslant \min\{y\cdot Z_\lambda/Z,y-\epsilon(\theta)\}\geqslant y/2$. 
Therefore, by Lemma~\ref{lem:gF}, it follows that
\begin{align*}
    \dfrac{\rmd F^{-1}(z|\theta)}{\rmd z}\bigg|_{z=\xi}\leqslant \dfrac{Ze^u}{[z^{1/p}(pZe^l)^{1/p}]^{p-1}}\bigg|_{z=\xi}\leqslant \dfrac{2^{(p-1)/p}Ze^u}{(pZe^l)^{(p-1)/p}}\cdot y^{1/p-1}
\end{align*}
Substitute it into the upper bound of $\phi_2(t)$, then it holds that
\begin{align*}
    I_2(\theta)\leqslant \lambda^q|2C_\epsilon(p)|^q\left(\dfrac{2Z^{1/p}e^u}{(pe^l)^{(p-1)/p}}\right)^{q}\int_{2C_\epsilon(p)\lambda}^{F_\lambda(R(\theta)|\theta)}y^{q/p-q}\rmd y\,.
\end{align*}
If $1/p+1/q>1$, then $p$ or $q$ equals $1$,
\begin{align*}
    I_2(\theta)&\leqslant \lambda^q|2C_\epsilon(p)|^q \left(\dfrac{2Z^{1/p}e^u}{(pe^l)^{(p-1)/p}}\right)^q \dfrac{F(R(\theta)|\theta)^{q(1/p+1/q-1)}}{q(1/p+1/q-1)}\\
    &\leqslant \lambda^q|2C_\epsilon(p)|^qp \left(\dfrac{2Z^{1/p}e^u}{(pe^l)^{(p-1)/p}}\right)^q \left(\dfrac{e^{-l}R^p}{pZ}\right)^{q(1/p+1/q-1)}\\
    &\leqslant\lambda^q|2C_\epsilon(p)|^q2^qe^{qu-l}Z^{q-1}R^{p+q-pq}\\
    &\leqslant \lambda^q|2C_\epsilon(p)|^q(2R)^q e^{q\omega(f,\K)+(q-1)/6}\times\dfrac{(2\pi e)^{p(q-1)/2}}{(\sqrt{\pi}p^{(p+1)/2})^{q-1}}\\
    &\leqslant \lambda^qp^{(p+1)/2}\left(\dfrac{20\sqrt{2\pi e}Re^{3\omega(f,\K)}a_1^{(p+1)/2}}{c_1^{1/2}}\right)^q\,.
  \end{align*}
If $1/p+1/q=1$, then $p=q=2$, 
\begin{align*}
    I_2(\theta)&\leqslant \lambda^q|2C_\epsilon(p)|^q\left(\dfrac{2Z^{1/p}e^u}{(pe^l)^{(p-1)/p}}\right)^q\left[\log(F(R(\theta)|\theta))-\log(2C_\epsilon(p)\lambda)\right]\\
    &\leqslant \lambda^q\left(\dfrac{4C_\epsilon(p)Z^{1/p}e^u}{(pe^l)^{(p-1)/p}}\right)^q\left[\log\left(\frac{1}{\lambda}\right)-\log\left(\frac{R}{r}\right)-\frac{p}{2}\log(2\pi e)-\log\left(\frac{10e^{\omega(f,\K)}}{\sqrt{2\pi}c_1^{1/2}}\right)\right]\\
    &\leqslant \lambda^q\log\left(\frac{1}{\lambda}\right)\cdot p^{(p-1)^2q/(2p)}\left(\frac{20e^{3\omega(f,\K)+1/(6p)}R\sqrt{2\pi e}}{\sqrt{\pi}^{1/p}c_1^{1/2}}\right)^q\\
    &\leqslant \lambda^q\log\left(\frac{1}{\lambda}\right)\cdot p^{1/2}\cdot \left(\frac{70e^{3\omega(f,\K)}R}{c_1^{1/2}}\right)^q
\end{align*}
If $1/p+1/q<1$, then
\begin{align*}
    I_2(\theta)&\leqslant \lambda^q|2C_\epsilon(p)|^q \left(\dfrac{2Z^{1/p}e^u}{(pe^l)^{(p-1)/p}}\right)^q \dfrac{(C_\epsilon(p)\lambda)^{q(1/p+1/q-1)}}{q(1-1/p-1/q)}\\
    &\leqslant \lambda^{(p+q)/p}\cdot\dfrac{2^q}{q(1-1/p-1/q)}\left[C_\epsilon(p)^{1/p+1/q}\cdot \dfrac{2Z^{1/p}e^u}{(pe^l)^{(p-1)/p}}\right]^q\\
    &\leqslant \lambda^{(p+q)/p}p^{\frac{(p+1)q}{2p}-\frac{(p-1)q}{p}}\cdot\dfrac{6}{q}\left[\left(\dfrac{5e^{2\omega(f,\K)}a_1^{(p+1)/2}}{c_1^{1/2}}\right)^{1/p+1/q}\cdot\dfrac{4e^{\omega(f,\K)+1/(6p)}R\sqrt{2\pi e}}{\sqrt{\pi}^{1/p}}\right]^q\\
    &\leqslant \lambda^{(p+q)/p}p^{\frac{(p+1)q}{2p}-\frac{(p-1)q}{p}}\cdot \frac{6}{q}\left(\frac{5^{1/p}e^{2\omega(f,\K)/p}a_1^{(p+1)/(2p)}}{c_1^{1/(2p)}}\right)^{p+q}\left(12e^{\omega(f,\K)}R\right)^q\\
    &\leqslant \lambda^{(p+q)/p}p^{\frac{(3-p)q}{2p}}\cdot\left(\dfrac{34 e^{\omega(f,\K)+2\omega(f,\K)/p}a_1^{(p+1)/(2p)}R^{q/(p+q)}}{c_1^{1/(2p)}}\right)^{p+q}
\end{align*}
Therefore, we obtain that
\begin{align*}
    \int_{\mathbb S_{p-1}}\phi_2(\theta)\rmd \theta&\leqslant \lambda^{(p+q)/p}(2^{2/p}a_2)^{p+q}p^{(p+q)/p}+ \omega_{p-1}\sup_{\theta\in\mathbb S_{p-1}}I_2(\theta)\\
    &\leqslant\lambda^{(p+q)/p}(2^{2/p}a_2))^{p+q}p^{(p+q)/p}+\dfrac{(2\pi e)^{p/2}}{\sqrt{\pi}}p^{-(p-1)/2}\sup_{\theta\in\mathbb S_{p-1}}I_2(\theta)\\
    &\leqslant (2^{2/p}a_2)^{p+q}p^{(p+q)/p}\lambda^{(p+q)/p)}+(5e^{\omega(f,\K)}a_2)^{p+q}p^{(p+q)/p}\lambda^{q\alpha}\log(1/\lambda)^{\mathbbm 1\{p=q=2\}}
\end{align*}
where $\alpha=\min(1/p+1/q,1)$. Here we omit the logarithmic factor of $\lambda$.
Therefore, we arrive at
\begin{align*}
    \int_{\mathbb S_{p-1}}\phi_2(\theta)\,\rmd \theta\leqslant (7e^{\omega(f,\K)}a_2)^{p+q}p^{(p+q)/p}\cdot\begin{cases}
        \lambda^q, &1/p+1/q>1\\
        \lambda^q\log(\frac{1}{\lambda}), &1/p+1/q=1\\
        \lambda^{q(1/p+1/q)}, &1/p+1/q<1
    \end{cases}
\end{align*}
\noindent\textbf{Step 3.} 
To establish the upper bound for $\phi_3$, we need the following lemma.
\begin{lemma}
\label{lem:Kt}
    Define the set $\K_t:=\{x\in\rp:\normtwo{x-\proj_{\K}(x)}\leqslant t\}.$ It holds for any $t>0$ that $\K_t\subset (1+t/r)\K$. Moreover, under Assumption~\ref{asm:dK}, it implies that 
    \begin{align*}
        \normtwo{x-\proj_{\K}(x)}^2\geqslant r^2(g_{\K}(x)-1)^2
    \end{align*}
    where $g_{\K}$ is the variation of Gauge function defined in Example~\ref{ex:gauge}.
\end{lemma}
By Lemma~\ref{lem:Kt}, it follows that
\[
\K_t\subset (1+\frac{t}{r})\K \,.
\]
In particular, this implies that for any $x\in\R^p,$
\[
x\in\left(1+\frac{\normtwo{x-\proj_{\K}(x)}}{r}\right)\K\,.
\]
Therefore, for any $x=t\theta\in\K^c$, we have
\begin{align*}
    t-R(\theta)\leqslant \left(1+\dfrac{\normtwo{x-\proj_{\K}(x)}}{r}\right)R(\theta)-R(\theta)=\dfrac{\normtwo{x-\proj_{\K}(x)}}{r}R(\theta)
\end{align*}
Using the inequality that $(a+b)^q\leqslant 2^{q-1}(a^q+b^q)$ and Assumption~\ref{asm:dK}, we have
\begin{align*}
    \phi_3(\theta)&=\int_{R(\theta)}^\infty \big|t-g(F_\lambda(t,\theta)-\epsilon(\theta),\theta\big|^q\dfrac{e^{-f(t\theta)-\frac{1}{2\lambda^2}d_{\K}(t\theta)}}{Z_\lambda}\rmd t\\
    &\leqslant \int_{R(\theta)}^\infty \big|t-R(\theta)+R(\theta)-g(F_\lambda(t,\theta)-\epsilon(\theta),\theta)\big|^q\dfrac{e^{-f(t\theta)-\frac{c_1}{2\lambda^2}\normtwo{t\theta-\proj_{\K}(t\theta)}^2}}{Z_\lambda}\rmd t\\
    &\leqslant \int_{R(\theta)}^\infty \big|t-R(\theta)+R(\theta)-g(F_\lambda(t,\theta)-\epsilon(\theta),\theta)\big|^q\dfrac{e^{-f(t\theta)-\frac{c_1r^2}{2\lambda^2R(\theta)^2}(t-R(\theta))^2}}{Z_\lambda}\rmd t
\end{align*}
Recall the definition of $\epsilon(\theta)$, and observe that for any $t\geqslant R(\theta)$,
\begin{align*}
    R(\theta)-g(F_\lambda(t,\theta)-\epsilon(\theta),\theta)&\leqslant R(\theta)-g(F_\lambda(R(\theta),\theta)-\epsilon(\theta),\theta)\\
    &=F^{-1}(F(R(\theta)|\theta)|\theta)-F^{-1}(F(R(\theta)|\theta)-\delta(\theta)|\theta)\\
    &= \delta(\theta)\cdot\dfrac{\rmd F^{-1}(z)}{\rmd z}\bigg|_{z=\xi}
\end{align*}
where $\delta(\theta)=F_\lambda(\infty,\theta)-F_\lambda(R(\theta),\theta),$ then following the proof of Lemma~\ref{lem:epsilon}, we have
\begin{align*}
    \delta(\theta)\leqslant \dfrac{1}{Z}\int_{R(\theta)}^\infty t^{p-1}e^{-f(t\theta)-\frac{1}{2\lambda^2}d_{\K}(t\theta)}\rmd t\leqslant \dfrac{2e^{-l}}{Zc_1^{1/2}}\cdot\dfrac{R^p}{r}\cdot\lambda\,.%1.531
\end{align*}
In addition,
\begin{align*}
    \dfrac{\rmd F^{-1}(z)}{\rmd z}\bigg|_{z=\xi}&\leqslant \dfrac{Ze^u}{[z^{1/p}(pZe^l)^{1/p}]^{p-1}}\bigg|_{z=\xi}\\
    &\leqslant \dfrac{2Ze^u}{F(R(\theta)|\theta)^{(p-1)/p}(pZe^l)^{(p-1)/p}}\\
    &\leqslant \dfrac{2Ze^ue^{(p-1)\omega(f,\K)/p}}{r^{p-1}}\,.
\end{align*}
Therefore 
\begin{align*}
    R(\theta)-g(F_\lambda(t,\theta)-\epsilon(\theta),\theta)\leqslant \dfrac{4e^{2\omega(f,\K)}}{c_1^{1/2}}\cdot\dfrac{R^p}{r^p}\lambda =a_3^p\lambda\,,
\end{align*}
where $a_3=\frac{4^{1/p}e^{2\omega(f,\K)/p}}{c_1^{1/(2p)}}\frac{R}{r}.$
Thus, we obtain that
\begin{align*}
    \phi_3(\theta)&\leqslant e^{-l}\int_{R(\theta)}^\infty|t-R(\theta)+a_3^p\lambda|^q\dfrac{e^{-\frac{c_1r^2}{2\lambda^2R(\theta)^2}(t-R(\theta))^2}}{Z}\rmd t\\
    &\leqslant \dfrac{e^{-l}\lambda}{Z}\int_0^\infty|\lambda s+a_3^p\lambda|^qe^{-\frac{c_1r^2}{2R(\theta)^2}s^2}\rmd s\\
    &=\dfrac{e^{-l}}{Z}\cdot \lambda^{q+1}\int_0^\infty|s+a_3^p|^qe^{-\frac{c_1r^2}{2R(\theta)^2}s^2}\rmd s\\
    &\leqslant \dfrac{\sqrt{2\pi}e^{\omega(f,\K)}p^{(p+1)/2}\lambda^{q+1}}{r^p(2\pi e)^{p/2}
    }\times 2^{q-1}\left(\dfrac{1}{2}\dfrac{(2R^2)^{(q+1)/2}}{(c_1r^2)^{(q+1)/2}}\Gamma(\dfrac{q+1}{2})+\dfrac{1}{2}a_3^{pq}\sqrt{\dfrac{2\pi R^2}{c_1r^2}}\right)\\
    &\leqslant \dfrac{2^{3q+2}\pi^{3/2}}{r^p
     }\cdot\frac{(2\pi e)^{p(q-1)/2}e^{(2q+1)\omega(f,\K)}}{c_1^{(q+1)/2}}\dfrac{R^{pq+1}}{r^{pq+1}}\cdot \lambda^{q+1}p^{(p+1)/2}
    \end{align*}
Thus,
\begin{align*}
    \int_{\mathbb S_{p-1}}\phi_3(\theta)\rmd\theta&\leqslant \omega_{p-1}\sup_{\theta\in \mathbb S_{p-1}}\phi_3(\theta)\\
    &\leqslant 2^{3q+2}\pi\cdot \dfrac{(2\pi e)^{pq/2}e^{(2q+1)\omega(f,\K)}}{c_1^{(q+1)/2}}\dfrac{R^{pq+1}}{r^{pq+p+1}}\cdot\lambda^{q+1}p\\
    &\leqslant a_4^{q+1}\dfrac{R^{pq+1}}{r^{pq+p+1}}\cdot\lambda^{q+1}p
\end{align*}
where $a_4=\frac{8(2\pi e)^{p/2}e^{2\omega(f,\K)}}{c_1^{1/2}}$.

Combining all the upper bounds of $\phi_i(\theta),i=1,2,3$, we find
\begin{align*}
    \wsq^q(\nu^\lambda,T_{\#}(\nu^\lambda))&\leqslant (2^{1/p}a_2)^{p+q}p^{(p+q)/p}\lambda^{(p+q)/p}\\
    &+(7e^{\omega(f,\K)}a_2)^{p+q}p^{(p+q)/p}\begin{cases}
        \lambda^q, &1/p+1/q>1\\
        \lambda^q\log(\frac{1}{\lambda}), &1/p+1/q=1\\
        \lambda^{q(1/p+1/q)}, &1/p+1/q<1\,.
   \end{cases}\\
    &+a_4^{q+1}\dfrac{R^{pq+1}}{r^{pq+p+1}}\cdot\lambda^{q+1}p
\end{align*}
Then we have 
\begin{align*}
    \wsq^q(\nu^\lambda,T_{\#}(\nu^\lambda))\lesssim a_2^{p+q}p^{(p+q)/p}\cdot\begin{cases}
        \lambda^q, &1/p+1/q>1\\
        \lambda^q\log(\frac{1}{\lambda}), &1/p+1/q=1\\
        \lambda^{q(1/p+1/q)}, &1/p+1/q<1\,.
   \end{cases}
\end{align*}
Recalling that $\wsq^q(\nu,T_{\#}(\nu^\lambda))\leqslant a_2^{p+q}p^{(p+q)/p}\lambda^{(p+q)/p}$, we arrive at the desired result
\begin{align*}
    \wsq(\nu,\nu^\lambda)\lesssim C(p,q)\cdot\begin{cases}
        \lambda, &1/p+1/q>1\\
        \lambda\log(\frac{1}{\lambda})^{1/q}, &1/p+1/q=1\\
        \lambda^{1/p+1/q}, &1/p+1/q<1\,.
    \end{cases}
\end{align*}
where $C(p,q)=\left[\frac{e^{4\omega(f,\K)}}{c_1^{1/2}}\cdot\left(\frac{\max(R,1)}{\min(r,1)}\right)^{(2p+1)}\cdot p\right]^{1/p+1/q}.$

\end{proof}

\subsection{Proof of Proposition~\ref{prop:lb}}
Assume $f$ is $M$-smooth and $d_{\K}$ is $M_0$-smooth. To prove Proposition~\ref{prop:lb}, we need the following auxiliary lemma.

\begin{lemma}
\label{lem:lb}
Under the assumptions stated in Proposition~\ref{prop:lb}, for any $k\geqslant 1$, when $\lambda=o(r/p)$, it holds that
\begin{align*}
\int_{\K^c}\normtwo{x}^k\nu^{\lambda}(\rmd x)/\mu_k(\nu)\geqslant \dfrac{\tilde C(c_1,c_2,M_0,M,\K)r^{p+k-1} (k+p)}{2\omega_{p-1}R^{k+p}}\lambda\,.
\end{align*}
where $\tilde C(c_1,c_2,M_0,M,\K)$ is a constant independent of $r$ and $p,k$.
\end{lemma}

We are now ready to prove Proposition~\ref{prop:lb}.

\begin{proof}[Proof of Proposition~\ref{prop:lb}]
Define
\begin{align*}
    \rho_1:=\dfrac{\int_{\K}\normtwo{x}^qe^{-f(x)}\rmd x}{\int_{\K}e^{-f(x)}\rmd x},\quad \rho_2(\lambda):=\dfrac{\int_{\K^c}\normtwo{x}^qe^{-f(x)-\frac{1}{2\lambda^2}d_{\K}(x)}\rmd x}{\int_{\K^c}e^{-f(x)-\frac{1}{2\lambda^2}d_{\K}(x)}\rmd x.}
\end{align*}
Then 
\begin{align*}
    \int_{\rp}\normtwo{x}^q\nu^{\lambda}(\rmd x)=\dfrac{\rho_1\int_{\K}e^{-f(x)}\rmd x+\rho_2(\lambda)\int_{\K^c}e^{-f(x)-\frac{1}{2\lambda^2}d_{\K}(x)}\rmd x}{\int_{\K}e^{-f(x)}\rmd x+\int_{\K^c}e^{-f(x)-\frac{1}{2\lambda^2}d_{\K}(x)}\rmd x},\quad \int_{\rp}\normtwo{x}^q\nu(\rmd x)=\rho_1\,.
\end{align*}
By Proposition 7.29 in~\cite{villani2009optimal}, it holds for any $q\geqslant 1$ that 
\begin{align*}
\wsq(\nu^\lambda,\nu)
\geqslant
\Big|\Big( \int_{\rp} \normtwo{x}^q \nu^{\lambda}(\rmd x)\Big)^{1/q}-\Big(\int_{\rp} \normtwo{x}^q \nu(\rmd x)\Big)^{1/q}\Big|\,.
\end{align*}
Note that when $|\delta|<\frac{1}{3}$, it holds that $|(1+\delta)^{1/k}-1|> \frac{1}{2k}|\delta|.$ Therefore, if we denote
\begin{align*}
    \delta:=\dfrac{\int_{\rp}\normtwo{x}^q\nu^{\lambda}(\rmd x)}{\int_{\rp}\normtwo{x}^q\nu(\rmd x)}-1\,,
\end{align*}
then we obtain that
\begin{align*}
    \wsq(\nu^\lambda,\nu)&\geqslant \left(\int_{\rp}\normtwo{x}^q\nu(\rmd x)\right)^{1/q}|\delta|\\
    &=\dfrac{1}{2q}\left(\int_{\rp}\normtwo{x}^q\nu(\rmd x)\right)^{1/q-1}\left|\int_{\rp}\normtwo{x}^q\nu^{\lambda}(\rmd x)-\int_{\rp}\normtwo{x}^q\nu(\rmd x)\right|\,.
\end{align*}
Therefore, it is enough to show the last term is $\Omega(\lambda).$
\begin{align}\label{eq:wsqerror}
    \left|\int_{\rp}\normtwo{x}^q\nu^{\lambda}(\rmd x)-\int_{\rp}\normtwo{x}^q\nu(\rmd x)\right|=\dfrac{\int_{\K^c}e^{-f(x)-\frac{1}{2\lambda^2}d_{\K}(x)}\rmd x}{\int_{\K}e^{-f(x)}\rmd x+\int_{\K^c}e^{-f(x)-\frac{1}{2\lambda^2}d_{\K}(x)}\rmd x}\cdot|\rho_2(\lambda)-\rho_1|\,.
\end{align}
By Lemma~\ref{lem:lb}, the first term on the right side is $\Omega(\lambda)$, then it remains to show $|\rho_2(\lambda)-\rho_1|=\Omega(1).$ To prove this, we derive the limit of $\rho_2(\lambda)$ when $\lambda$ approaches zero, and the desired result follows by the condition.

For Case 1 with $Q=I$, $d_{\K}(x)=\normtwo{x-\proj_{\K}(x)}^2=\text{dist}(x,\K)^2$. We assume that $\K$ has a $C^2$-boundary, then it holds for any $\varepsilon>0$ that 
\begin{align*}
    &\quad\lim_{\lambda\to 0}\dfrac{1}{\lambda}\int_{\K^c}\normtwo{x}^qe^{-f(x)-\frac{1}{2\lambda^2}d_{\K}(x)}\rmd x\\
    &=\lim_{\lambda\to 0}\dfrac{1}{\lambda}\int_{\bar{\K}_\varepsilon-\K}\normtwo{x}^qe^{-f(x)-\frac{1}{2\lambda^2}\text{dist}(x,\K)^2}\rmd x+\lim_{\lambda\to 0}\dfrac{1}{\lambda}\int_{\bar{\K}_\varepsilon^c}\normtwo{x}^qe^{-f(x)-\frac{1}{2\lambda^2}\text{dist}(x,\K)^2}\rmd x\\
    &=\lim_{\lambda\to 0}\dfrac{1}{\lambda}\int_{\bar{\K}_\varepsilon-\K}\normtwo{x}^qe^{-f(x)-\frac{1}{2\lambda^2}\text{dist}(x,\K)^2}\rmd x\,.
\end{align*}
where $\bar{\K}_{\varepsilon}:=\{x\in\rp:\text{dist}(x,\K)\leqslant\varepsilon\}$, the last equality follows from Dominated Convergence Theorem (DCT). To evaluate the integral $I=\int_{\bar{\K}_\varepsilon-\K}\normtwo{x}^qe^{-f(x)-\frac{1}{2\lambda^2}\text{dist}(x,\K)^2}\rmd x$, we utilize a tubular neighborhood parameterization. The diffeomorphism $\Phi:\partial\K\times(0,\varepsilon)\to\K^c$, defined as $x=z+tv(z)$, uniquely maps points near $\partial \K$ to coordinates $(z,t)$, where $z\in\partial\K$ and $t=\text{dist}(x,\K)$. The Jacobian determinant $J(z,t)$, accounting for curvature effects, is $\prod_{i=1}^{p-1}(1-t\kappa_i(z))$, with $\kappa_i(z)$ as principal curvatures. Thus
\begin{align*}
    \lim_{\lambda\to 0}\dfrac{1}{\lambda}I&=\lim_{\lambda\to 0}\dfrac{1}{\lambda}\int_{\partial\K}\int_0^{\varepsilon}\normtwo{z+tv(z)}^qe^{-f(z+tv(z))}J(z,t)e^{-\frac{t^2}{2\lambda^2}}\rmd t\,\rmd\sigma(z)\\
    &=\lim_{\lambda\to 0}\dfrac{1}{\lambda}\int_{\partial\K}\int_0^{\infty}\normtwo{z+tv(z)}^qe^{-f(z+tv(z))}J(z,t)e^{-\frac{t^2}{2\lambda^2}}\rmd t\,\rmd\sigma(z)\,.
\end{align*}
where $J(z,t):=1$ if $t>\varepsilon$, the second equality follows from DCT. After parameterizing $x=z+tv(z)$, a substitution for $t$ can further simplify the radial integral. Defining $s=t/\lambda$, the integral becomes
\begin{align*}
    \lim_{\lambda\to 0}\dfrac{1}{\lambda}I&=\lim_{\lambda\to 0}\int_{\partial\K}\int_0^{\infty}\normtwo{z+\lambda sv(z)}^qe^{-f(z+\lambda sv(z))}J(z,\lambda s)e^{-\frac{s^2}{2}}\rmd s\,\rmd\sigma(z)\\
    &=\int_{\partial\K}\int_0^{\infty}\normtwo{z}^qe^{-f(z)}e^{-\frac{s^2}{2}}\rmd s\,\rmd\sigma(z)\\
    &=\sqrt{\dfrac{\pi}{2}}\int_{\partial\K}\normtwo{z}^qe^{-f(z)}\rmd\sigma(z)\,.
\end{align*}
Similarly, we obtain that
\begin{align*}
    \lim_{\lambda\to 0}\dfrac{1}{\lambda}\int_{\K^c}e^{-f(x)-\frac{1}{2\lambda^2}d_{\K}(x)}\rmd x=\sqrt{\dfrac{\pi}{2}}\int_{\partial\K}e^{-f(z)}\rmd \sigma(z)\,.
\end{align*}
Therefore, 
\begin{align*}
    \lim_{\lambda\to 0}\rho_2(\lambda)=\dfrac{\int_{\partial \K}\normtwo{z}^qe^{-f(z)}\rmd\sigma(z)}{\int_{\partial\K}e^{-f(z)}\rmd\sigma(z)}\,.
\end{align*}
For Case 2, we employ a parameterization aligned with the geometric interpretation of the Gauge function. For $x\in\K^c$, the Gauge function $\gamma_{\K}(x)$ defines the minimal scaling factor $s\geqslant 1$ such that $x=sz$ for some $z\in\partial\K$. This allows the parameterization $x=(1+t)z$, where $t>0$ and $z\in\partial\K$, ensuring $\gamma_{\K}(x)=1+t$. The volume element $\rmd x$ transforms under this radial scaling as $\rmd x=(1+t)^{p-1}\rmd\sigma(z)\,\rmd t$, therefore,
\begin{align*}
    &\quad\lim_{\lambda\to 0}\dfrac{1}{\lambda}\int_{\K^c}\normtwo{x}^qe^{-f(x)-\frac{1}{2\lambda^2}d_{\K}(x)}\rmd x\\
    &=\lim_{\lambda\to 0}\dfrac{1}{\lambda}\int_{\partial \K}\int_0^{\infty}\normtwo{(1+t)z}^qe^{-f((1+t)z)}e^{-\frac{t^2}{2\lambda^2}}(1+t)^{p-1}\rmd t\,\rmd\sigma(z)\\
    &=\lim_{\lambda\to 0}\int_{\partial \K}\int_0^{\infty}\normtwo{(1+\lambda s)z}^ke^{-f((1+\lambda s)z)}e^{-\frac{s^2}{2}}(1+\lambda s)^{p-1}\rmd s\,\rmd\sigma(z)\\
    &=\sqrt{\dfrac{\pi}{2}}\int_{\partial\K}\normtwo{z}^qe^{-f(z)}\rmd\sigma(z)\,.
\end{align*}
The last equality follows from DCT, and similarly we obtain that
\begin{align*}
    \lim_{\lambda\to 0}\rho_2(\lambda)=\dfrac{\int_{\partial \K}\normtwo{z}^qe^{-f(z)}\rmd\sigma(z)}{\int_{\partial\K}e^{-f(z)}\rmd\sigma(z)}\,.
\end{align*}
which is the same as Case 1.
\end{proof}

\subsection{Proof of Auxiliary Lemmas}

\begin{proof}[Proof of Lemma~\ref{lem:ext_int}]
By Assumption~\ref{asm:dK}, we have
\begin{align*}
    \int_{\K^c}\|x\|_2^ke^{-U^{\lambda}(x)}\rmd x=\int_{\K^c}\|x\|_2^ke^{-f(x)}\cdot e^{-\frac{1}{2\lambda^2}d_{\K}(x)}\rmd x\leqslant \int_{\K^c}\|x\|_2^ke^{-f(x)}\cdot e^{-\frac{c_1}{2\lambda^2}\normtwo{x-\proj_{\K}(x)}^2}\,.
\end{align*}
By Fubini's theorem and the fact that
\begin{align*}
\int_0^{\infty} \frac{c_1 t}{\lambda^2} e^{-\frac{c_1t^2}{2\lambda^2}}\1\{t\geqslant\normtwo{x-\proj_{\K}(x)}\}\rmd t =  e^{- \frac{c_1}{2\lambda^2} \,\normtwo{x-\proj_{\K}(x)}^2}\,,
\end{align*}
it holds for every $k\geqslant 0$ that
\begin{align}
\label{psi:1}  
    \int_{\K^c}\normtwo{x}^k e^{-U^\lambda(x)} \rmd x
    \leqslant\int_0^{\infty} \Big(\underbrace{\int_{\K^c}\normtwo{x}^k 
    e^{-f(x)} \1\{\,\normtwo{x-\proj_{\K}(x)} 
    \leqslant t\}\, \rmd x}_{\psi(t)}\,\Big) \frac{c_1 t}{\lambda^2} e^{-\frac{c_1t^2}{2\lambda^2}} \,\rmd t .
\end{align}
Then we focus on $\psi(t)$, that is
\begin{align*}
    \psi(t)&\leqslant e^{-l}\int_{\K^c}\normtwo{x}^k\1\{\,\normtwo{x-\proj_{\K}(x)}\leqslant t\}\,\rmd x\\
    &=e^{-l}\left(\int_{\K_t}\normtwo{x}^k\,\rmd x-\int_{\K}\normtwo{x}^k\,\rmd x\right)\\
    &\leqslant e^{-l}\left(\int_{(1+t/r)\K}\normtwo{x}^k\,\rmd x-\int_{\K}\normtwo{x}^k\,\rmd x\right)\,.
\end{align*}
where $l$ is the lower bound of $f$ in $\rp$. The last inequality use the fact that $\K_t\subset(1+t/r)\K$ proved in Lemma~\ref{lem:Kt}. Therefore we have
\begin{align*}
    \psi(t)\leqslant e^{-l}\left[\left(1+t/r\right)^{p+k}-1\right]\int_{\K}\normtwo{x}^k\,\rmd x\leqslant e^{-l}\cdot \left(e^{(p+k)t/r}-1\right)\int_{\K}\normtwo{x}^k\rmd x\,.
\end{align*}
Substituting this upper bound in \eqref{psi:1} leads to
\begin{align*}
    \int_{\K^c}\normtwo{x}^ke^{-U^\lambda(x)}\rmd x&\leqslant \underbrace{\int_0^{\infty}\dfrac{c_1 t}{\lambda^2}e^{-\frac{c_1t^2}{2\lambda^2}}(e^{(p+k)t/r}-1)\,\rmd t}_{\text I}\times e^{-l}\int_{\K}\normtwo{x}^k\rmd x
\end{align*}
Next, we derive a closed-form upper bound for the integral term,
\begin{align*}
    \text{I}&=\int_0^{\infty}  \frac{c_1 t}{\lambda^2} e^{-\frac{c_1t^2}{2\lambda^2}}  \big( e^{(p+k)t/r}-1\big)\,\rmd t\\
    &=\int_0^{\infty} \big( e^{(p+k)t/r}-1\big)\rmd\left(-e^{-c_1t^2/2\lambda^2}\right)\\
    &=\dfrac{p+k}{r}\int_0^{\infty}\exp\left\{-\frac{c_1t^2}{2\lambda^2}+\frac{(p+k)t}{r}\right\}\,\rmd t\,.
\end{align*}
By Young's inequality, we have
\begin{align*}
    \dfrac{(p+k)t}{r}\leqslant\dfrac{\eta t^2}{2}+\dfrac{(p+k)^2}{2r^2\eta},\quad \eta>0
\end{align*}
Setting $\eta=\frac{c_1}{2\lambda^2}$ and combining this with the previous displays, we obtain
\begin{align*}
    \text{I}&\leqslant\dfrac{p+k}{r}\exp\left(\dfrac{\lambda^2(p+k)^2}{c_1r^2}\right)\int_0^{\infty}\exp\left(-\dfrac{c_1t^2}{4\lambda^2}\right)\rmd t\\
    &\leqslant\dfrac{(p+k)\Gamma(3/2)}{r}\left(\dfrac{4\lambda^2}{c_1}\right)^{1/2}\exp\left(\dfrac{\lambda^2(p+k)^2}{c_1r^2}\right)\,.
\end{align*}
When $\lambda<c_1^{1/2}r/(p+k)$, it follows that
\begin{equation}
\begin{aligned}
    \int_{\K^c}\normtwo{x}^ke^{-U^\lambda(x)}\rmd x&\leqslant \dfrac{3\sqrt{\pi}(p+k)}{rc_1^{1/2}}\lambda\times e^{-l}\int_{\K}\normtwo{x}^k\,\rmd x\,.
\end{aligned}
\label{eq:Zlambda}
\end{equation}
\end{proof}

\begin{proof}[Proof of Lemma~\ref{lem:epsilon}]
    By rearranging the term and the triangle inequality, we have
\begin{align*}
    |\epsilon(\theta)|&=\left|\dfrac{\int_0^{R(\theta)}t^{p-1}e^{-f(t\theta)}\rmd t+\int_{R(\theta)}^\infty t^{p-1}e^{-f(t\theta)-\frac{1}{2\lambda^2}d_{\K}(t\theta)}\rmd t}{Z_\lambda}-\dfrac{\int_0^{R(\theta)}t^{p-1}e^{-f(t\theta)}\rmd t}{Z}\right|\\
    &\leqslant \left|\dfrac{1}{Z_\lambda}-\dfrac{1}{Z}\right| \cdot\int_0^{R(\theta)}t^{p-1}e^{-f(t\theta)}\rmd t +\dfrac{1}{Z_\lambda}\int_{R(\theta)}^\infty t^{p-1}e^{-f(t\theta)-\frac{1}{2\lambda^2}d_{\K}(t\theta)}\rmd t\\
\end{align*}
Recall the display \eqref{eq:Zdiff}, then we have
\begin{align*}
    |\epsilon(\theta)|&\leqslant \int_0^{R(\theta)}t^{p-1}e^{-f(t\theta)}\rmd t\times \dfrac{Z_\lambda-Z}{Z^2}+\int_{R(\theta)}^\infty t^{p-1}e^{-f(t\theta)-\frac{1}{2\lambda^2}d_{\K}(t\theta)}\rmd t\times \dfrac{1}{Z}\\
    &\leqslant \int_0^Rt^{p-1}e^{-l}\rmd t\times\dfrac{3\sqrt{\pi}pe^{-l}\text{Vol}(\K)}{rc_1^{1/2}Z^2}\lambda+\dfrac{e^{-l}}{Z}\int_{R(\theta)}^\infty t^{p-1}e^{-\frac{c_1}{2\lambda^2}\normtwo{t\theta-\proj_{\K}(t\theta)}^2}\rmd t\\
    &\leqslant \dfrac{3\sqrt{\pi}R^pe^{-2l}\text{Vol}(\K)}{rc_1^{1/2}Z^2}\lambda+\dfrac{e^{-l}}{Z}\underbrace{\int_{R(\theta)}^{\infty} t^{p-1}e^{-\frac{c_1r^2}{2\lambda^2R(\theta)^2}(t-R(\theta))^2}\rmd t}_{I(\theta)}\,.
\end{align*}
The last inequality follows from Lemma~\ref{lem:Kt}, since $g_{\K}(t\theta)=[t-R(\theta)]/R(\theta)$. Substitute $s=[t-R(\theta)]/\lambda,$ 
\begin{align*}
    I(\theta)&=\lambda\times \int_0^\infty (R(\theta)+\lambda s)^{p-1}e^{-\frac{c_1r^2}{2R(\theta)^2}s^2}\rmd s\\
    &\leqslant\lambda\times\int_0^\infty (R+\lambda s)^{p-1}e^{-\frac{c_1r^2}{2R^2}s^2}\rmd s\\
    &\leqslant \lambda\times 2^{p-1}\left(\dfrac{\sqrt{\pi}R^{p-1}}{2\sqrt{c_1r^2/2R^2}}+\dfrac{\lambda^{p-1}\Gamma(p/2)} {2(c_1r^2/2R^2)^{p/2}}\right)
\end{align*}
Therefore, we have
\begin{align*}
    |\epsilon(\theta)|\leqslant \left[\dfrac{3\sqrt{\pi}R^pe^{-2l}\text{Vol}(\K)}{rc_1^{1/2}Z^2}+\frac{2^{p-1}e^{-l}}{Z}\left(\dfrac{\sqrt{\pi}R^{p-1}}{2\sqrt{c_1r^2/2R^2}}+\dfrac{\lambda^{p-1}\Gamma(p/2)}{2(c_1r^2/2R^2)^{p/2}}\right)\right]\lambda\,,
\end{align*}
where the coefficient is independent of $\theta.$ Note that $e^{-u}\text{Vol}(\K)\leqslant Z\leqslant e^{-l}\text{Vol}(\K)$ and $r^pV_p\leqslant\text{Vol}(\K)\leqslant R^pV_p$, and we have
\begin{align*}
    \dfrac{1}{\sqrt{2\pi}}\cdot \dfrac{(2\pi e)^{p/2}}{p^{(p+1)/2}}\leqslant V_p\leqslant \dfrac{e^{1/6}}{\sqrt{\pi}}\cdot\dfrac{(2\pi e)^{p/2}}{p^{(p+1)/2}},\quad \dfrac{2\sqrt{\pi}p^{(p-1)/2}}{(2e)^{p/2}}\leqslant\Gamma\left(\dfrac{p}{2}\right)\leqslant \dfrac{2\sqrt{\pi}e^{1/6}p^{(p-1)/2}}{(2e)^{p/2}}\,.
\end{align*}
Thus when $\lambda<c_1^{1/2}r/\sqrt{p}$, we obtain that 
\begin{align*}
    \dfrac{|\epsilon(\theta)|}{\lambda}&\leqslant\dfrac{3\sqrt{\pi}R^pe^{2\omega(f,\K)}}{rc_1^{1/2}r^p}\cdot  \dfrac{\sqrt{2\pi}p^{(p+1)/2}}{(2\pi e)^{p/2}}+\dfrac{\pi 2^{p-1}e^{\omega(f,\K)}p^{(p+1)/2}}{r^p(2\pi e)^{p/2}}\cdot\dfrac{R^p}{c_1^{1/2}r}\\
    &\quad+\dfrac{2^{p-1}e^{\omega(f,\K)}\sqrt{2\pi}p^{(p+1)/2}}{r^p(2\pi e)^{p/2}}\cdot\dfrac{\sqrt{\pi}e^{1/6}R^p}{e^{p/2}c_1^{1/2}r}\\
    &=\dfrac{3\sqrt{2}\pi e^{2\omega(f,\K)}}{c_1^{1/2}(2\pi e)^{p/2}}\cdot\dfrac{R^pp^{(p+1)/2}}{r^{p+1}}+\dfrac{2^{p-1}\pi e^{\omega(f,\K)}}{(2\pi e)^{p/2}c_1^{1/2}}\cdot\dfrac{R^pp^{(p+1)/2}}{r^{p+1}}\\
    &\quad+\dfrac{2^{p-1}e^{1/6}\sqrt{2}\pi e^{\omega(f,\K)}}{(2\pi e)^{p/2}e^{p/2}c_1^{1/2}}\cdot\dfrac{R^pp^{(p+1)/2}}{r^{p+1}}\\
    &\leqslant \dfrac{5e^{2\omega(f,\K)}}{c_1^{1/2}}\cdot\dfrac{R^pp^{(p+1)/2}}{r^{p+1}}\,.% 4.756
\end{align*}
Then let $a_1=\frac{\max (R^2,1)}{r^2}$, and $C_\epsilon(p)=\frac{5e^{2\omega(f,\K)}}{c_1^{1/2}}(a_1p)^{(p+1)/2}$, we have $|\epsilon(\theta)|\leqslant C_\epsilon(p)\lambda$.

\end{proof}

\begin{proof}[Proof of Lemma~\ref{lem:Kt}]
It holds for any $x\in\K_t/(1+b)$ that  
\begin{align*}
\normtwo{(1+b)x-\proj_{\K}\big((1+b)x\big)} \leqslant t
\end{align*}
which is equivalent to
\begin{align*}
\frac{1+b}{b}\,\normtwo{x-\frac{1}{1+b}\proj_{\K}\big((1+b)x\big)} \leqslant \frac{t}{b}\,.
\end{align*}
Note that we can rewrite $x$ as
\begin{align*}
x = \frac{1}{1+b}\proj_{\K}\big((1+b)x\big)
+\frac{b}{1+b}\frac{x-\frac{1}{1+b}\proj_{\K}\big((1+b)x\big)}{\frac{b}{1+b}}\,.
\end{align*}
Notice that
\begin{align*}
\proj_{\K}\big((1+b)x\big)\in\K
\end{align*}
and when $b= t/r$ we have
\begin{align*}
\frac{x-\frac{1}{1+b}\proj_{\K}\big((1+b)x\big)}{\frac{b}{1+b}}\in\K \,.
\end{align*}
By the convexity of the set $\K$, this implies $x\in\K,$ then we obtain that $\K_t\subset(1+\frac{t}{r})\K.$ For the second claim, recall the definition of $g_{\K}$, it holds for any $x\in\K_t$ that $g_{\K}(x)\leqslant 1+t/r.$ Thus set $t=\normtwo{x-\proj_{\K}(x)}$, then we have
\begin{align*}
    g_{\K}(x)\leqslant 1+\dfrac{\normtwo{x-\proj_{\K}(x)}}{r}
\end{align*}
which reach the desired result.
\end{proof}

\begin{proof}[Proof of Lemma~\ref{lem:lb}]
By the definition of $\nu^\lambda$ and $\mu_k(\nu),$ we have
\begin{align}
\label{eq:twoparts}
\int_{\K^c}\normtwo{x}^k\nu^\lambda(\rmd x)/\mu_k(\nu) 
& = \frac{\int_{\K^c}\normtwo{x}^k e^{-f(x) -\frac{1}{2\lambda^2}d_{\K}(x)}\rmd x}{\int_{\rp}e^{-U^\lambda(x)}\rmd x}\cdot
\frac{\int_{\K}e^{-f(x)}\rmd x}{\int_{\K}\normtwo{x}^k e^{-f(x)}\rmd x}\,.
\end{align}
Note that $f$ is $M$-smooth, it then holds that
\begin{align*}
 f(x)\leqslant f(0)+\frac{M}{2}\normtwo{x}^2\,.
\end{align*}
This implies
\begin{align*}
\int_{\K^c}\normtwo{x}^k e^{-f(x) -\frac{1}{2\lambda^2}d_{\K}(x)}\rmd x
\geqslant
e^{-f(0)}\int_{\K^c}\normtwo{x}^k e^{-\frac{M}{2}\normtwo{x}^2 -\frac{c_2}{2\lambda^2}\normtwo{x-\proj_{\K}(x)}^2}\rmd x\,.
\end{align*}
By Fubini's theorem and the fact that
\begin{align*}
    \int_0^{\infty}\dfrac{c_2 t}{\lambda^2}e^{-\frac{c_2t^2}{2\lambda^2}}\1_{[\,\normtwo{x-\proj_{\K}(x)},+\infty)}(t)\,\rmd t=e^{-\frac{c_2}{2\lambda^2}\normtwo{x-\proj_{\K}(x)}^2}\,,
\end{align*}
it holds for every $k\geqslant 0$ that
\begin{align*}
    \int_{\K^c}\normtwo{x}^ke^{-f(x)-\frac{1}{2\lambda^2}d_{\K}(x)}\rmd x\geqslant e^{-f(0)}\int_0^{\infty}\underbrace{\left(\int_{\K^c}\normtwo{x}^ke^{-\frac{M}{2}\normtwo{x}^2}\1_{[\,\normtwo{x-\proj_{\K}(x)},+\infty)}(t)\,\rmd x\right)}_{\psi(t)}\dfrac{c_2 t}{\lambda^2}e^{-\frac{c_2t^2}{2\lambda^2}}\,\rmd t.
\end{align*}
By Assumption~\ref{asm:dK}, $\normtwo{x-\proj_{\K}(x)}\leqslant c_1^{-1/2}d_{\K}(x)^{1/2}$, therefore 
\begin{align*}
    \psi(t)\geqslant\int_{\K^c}\normtwo{x}^ke^{-\frac{M}{2}\normtwo{x}^2}\1(d_{\K}(x)&\leqslant c_1t^2)\,\rmd x\,.
\end{align*}
Define the set $\tilde\K_t:=\{x\in\rp:d_{\K}(x)\leqslant c_1t^2\}$. Since $d_{\K}$ is $M_0$-smooth, then it holds for any $x\in\K$ that
\begin{align*}
    d_{\K}((1+b)x)\leqslant d_{\K}(x)+\langle\nabla d_{\K}(x),bx\rangle+M_0\,\normtwo{bx}^2=M_0b^2\,\normtwo{x}^2\,.
\end{align*}
Let $b=\sqrt{c_1/M_0}\,t/R$, then $d_{\K}((1+b)x)\leqslant c_1t^2$, it implies that $(1+b)\K\subseteq \tilde\K_t.$ We then obtain that
\begin{align*}
    \psi(t)&\geqslant \int_{\tilde\K_t}\normtwo{x}^ke^{-\frac{M}{2}\normtwo{x}^2}\,\rmd x-\int_{\K}\normtwo{x}^ke^{-\frac{M}{2}\normtwo{x}^2}\,\rmd x\\
    &\geqslant\int_{(1+b)\K}\normtwo{x}^ke^{-\frac{M}{2}\normtwo{x}^2}\,\rmd x-\int_{\K}\normtwo{x}^ke^{-\frac{M}{2}\normtwo{x}^2}\,\rmd x\\
    &=\int_{(1+b)\K-\K}\normtwo{x}^ke^{-\frac{M}{2}\normtwo{x}^2}\,\rmd x\,.
\end{align*}
Denote the right side of the last equality by $\psi_1(t)$, then $\psi(t)\geqslant\psi_1(t)$, we have
\begin{align*}
    \int_{\K^c}\normtwo{x}^ke^{-f(x)-\frac{1}{2\lambda^2}d_{\K}(x)}\rmd x&\geqslant e^{-f(0)}\int_0^{\infty}\psi_1(t)\dfrac{c_2t}{\lambda^2}e^{-\frac{c_2t^2}{2\lambda^2}}\rmd t\\
    &=e^{-f(0)}\int_0^{\infty}\psi_1(t)d\left(-e^{-c_2t^2/2\lambda^2}\right)\rmd t\\
    &=e^{-f(0)}\int_0^{\infty}\psi_1'(t)e^{-\frac{c_2t^2}{2\lambda^2}}\rmd t\,.
\end{align*}
To compute $\psi_1'(t)$, we apply the Reynolds Transport Theorem to the t-dependent domain $(1+b(t))\K-\K$, where $b(t)=\frac{\sqrt{c_1/M_0}}{R}t$. Parametrize the outer boundary $\partial[(1+b(t))\K]$ by $x=(1+b(t))y$ with $y\in \partial\K$. The outward normal velocity of the moving boundary is $v_{\vec n}(x)=\frac{\rmd b(t)}{\rmd t}\cdot \normtwo{\nabla\gamma_{\K}(x)}^{-1}$. By the theorem, the derivative becomes a boundary integral over $\partial[(1+b(t))\K]$:
\begin{align*}
    \psi_1'(t)=\int_{\partial[(1+b(t))\K]}\normtwo{x}^ke^{-\frac{M}{2}\normtwo{x}^2}v_{\vec n}(x)\,\rmd S(x)\,.
\end{align*}
Substitute $x=(1+b)y$ and $\rmd S(x)=(1+b)^{p-1}\rmd S(y)$, we obtain
\begin{align*}
    \psi_1'(t)&=\dfrac{\sqrt{c_1/M_0}}{R}(1+b(t))^{k+p-1}\int_{\partial\K}\dfrac{\normtwo{y}^k}{\normtwo{\nabla\gamma_{\K}(y)}}e^{-\frac{M}{2}(1+b)^2\normtwo{y}^2}\rmd S(y)\\
    &\geqslant \dfrac{\sqrt{c_1/M_0}r^{k+2}\mathcal H_{p-1}(\partial\K)}{R}(1+b(t))^{k+p-1}e^{-\frac{MR^2}{2}(1+b)^2}\\
    &\geqslant C(c_1,M_0,\K)r^{p+k+1}e^{-\frac{MR^2}{2}(1+b)^2}\,.
\end{align*}
where the constant $C(c_1,M_0,\K)$ is independent of $r,p$ and $k$. Therefore, it follows that
\begin{align*}
    \int_{\K^c}\normtwo{x}^ke^{-f(x)-\frac{1}{2\lambda^2}d_{\K}(x)}\rmd x&\geqslant C(c_1,M_0,\K)r^{p+k+1}\int_0^{\infty}e^{-\frac{c_2t^2}{2\lambda^2}-\frac{MR^2}{2}(1+at)^2}\rmd t\\
    &\geqslant \tilde C(c_1,c_2,M_0,M,\K)r^{p+k-1}\lambda\,.
\end{align*}
where $a=\frac{\sqrt{c_1/M_0}}{R}$. Moreover, by Lemma~\ref{lem:ext_int}, we find
\begin{align*}
    \int_{\rp}e^{-U^{\lambda}(x)}\rmd x&=\int_{\K}e^{-U^{\lambda}(x)}\rmd x+\int_{\K^c}e^{-U^{\lambda}(x)}\rmd x\\
    &\leqslant \left(1+\dfrac{3\sqrt{\pi}p\lambda}{rc_1^{1/2}}\right)\mu_0(\nu)\,.
\end{align*}
Furthermore, using the integration in polar coordinates gives
\begin{align*}
    \int_{\K}\normtwo{x}^ke^{-f(x)}\rmd x\leqslant \omega_{p-1}\int_0^Rt^{k+p-1}\rmd t=\omega_{p-1}\dfrac{R^{k+p}}{k+p}
\end{align*}
Collecting pieces and plugging into display~\eqref{eq:twoparts}
\begin{align*}
    \int_{\K^c}\normtwo{x}^k\nu^{\lambda}(\rmd x)/\mu_k(\nu)\geqslant \dfrac{\tilde C(c_1,c_2,M_0,M,\K)r^{p+k-1} (k+p)}{(1+3\sqrt{\pi}p\lambda/rc_1^{1/2})\omega_{p-1}R^{k+p}}\lambda\,.
\end{align*}
\end{proof}

\section{Proofs of Section~\ref{sec:rlmc} and Section~\ref{sec:rklmc}}
\label{app:exam}

\begin{proof}[Proof of Lemma~\ref{lem:scv-smooth}]
We first consider the Bregman projection.\\ 
\textbf{I. Bregman Projection}~~
We factorize the matrix $Q^{-1}$ as $Q^{-1}=DD^\top$.
Define the composition operator $\ell_{\K}\circ D(x)=\ell_{\K}(Dx).$
It then follows that
\begin{align*}
\ell_{\K}^{B,\lambda}(x)&=\inf_{y\in\mathbb R^p}\left(\ell_{\K}(y)+\dfrac{1}{2\lambda^2}(x-y)^\top Q(x-y)\right)\\
&=\inf_{y\in\mathbb R^p}\left(\ell_{\K}\circ D(y)+\dfrac{1}{2\lambda^2}(D^{-1}x-y)^\top D^\top QD(D^{-1}x-y)\right)\\
&=\inf_{y\in\mathbb R^p}\left(\ell_{\K}\circ D(y)+\dfrac{1}{2\lambda^2}\|D^{-1}x-y\|_2^2\right).
\end{align*}
Define $e_{\lambda}(\ell_{\K}\circ D)(x):=\ell_{\K}^{B,\lambda}\circ D(x)$, then 
\begin{align*}
e_{\lambda}(\ell_{\K}\circ D)(x)=\inf_{y\in\mathbb R^p}\left(\ell_{\K}\circ D(y)+\dfrac{1}{2\lambda^2}\|x-y\|_2^2\right)
\end{align*}
is a Moreau envelope function of the composite function $\ell_{\K}\circ D$, and the corresponding proximal mapping is 
\begin{align*}
P_{\lambda}(\ell_{\K}\circ D)(x):=\argmin_{y\in D^{-1}\K}\left(\ell_{\K}\circ D(y)+\dfrac{1}{2\lambda^2}\|x-y\|_2^2\right).
\end{align*}
By Theorem 2.26 in~\cite{rockafellar2009variational}, the envelope function $e_{\lambda}(\ell_{\K}\circ D)$ is convex, continuously differentiable, and its gradient is as follows
\begin{align*}
\nabla e_{\lambda}(\ell_{\K}\circ D)(x)=\dfrac{1}{\lambda^2}(x-P_{\lambda}(\ell_{\K}\circ D)(x)).
\end{align*}
Note that $\ell_{\K}\circ D$ is convex, then $P_{\lambda}(\ell_{\K}\circ D)$ is maximal monotone and nonexpansive. 
By Proposition 12.19 in~\cite{rockafellar2009variational}, we have
\begin{align*}
\|\nabla e_{\lambda}(\ell_{\K}\circ D)(x_1)-\nabla e_{\lambda}(\ell_{\K}\circ D)(x_2)\|_2\leqslant\dfrac{1}{\lambda^2}\|x_1-x_2\|_2.
\end{align*}
Recall that $e_{\lambda}\mathscr (l_{\K}\circ D)(x)=\ell_{\K}^{B,\lambda}\circ D(x)$, it then follows that
\begin{align*}
\|\nabla \ell_{\K}^{B,\lambda}(x_1)-\nabla\ell_{\K}^{B,\lambda}(x_2)\|_2&=\|D^{-\top}\nabla \mathscr (l_{\K}^{B,\lambda}\circ D)(D^{-1}x_1)-D^{-\top}\nabla(\ell_{\K}^{B,\lambda}\circ D)(D^{-1}x_2)\|_2\\
&=\|D^{-\top}\nabla e_{\lambda}(\ell_{\K}\circ D)(D^{-1}x_1)-D^{-\top}\nabla e_{\lambda}(\ell_{\K}\circ D)(D^{-1}x_2)\|_2\\
&\leqslant\dfrac{1}{\lambda^2}\|D^{-1}\|_2\|D^{-1}x_1-D^{-1}x_2\|_2\\
&\leqslant\dfrac{1}{\lambda^2}\|D^{-1}\|_2^2\|x_1-x_2\|_2.
\end{align*}
Since $Q^{-1}=DD^\top$, it holds that $\|D^{-1}\|_2^2=\lambda_{\max}(Q)$. 
Therefore, we obtain
\begin{align*}
M^{\lambda}=M+\dfrac{\lambda_{\max}(Q)}{\lambda^2}.
\end{align*}
Now we show that $\int_{\rp} e^{-U^{B,\lambda}(x)}\rmd x<\infty.$ 
Note that
\begin{align*}
\int_{\rp}e^{-U^{B,\lambda}(x)}\rmd x
=&\int_{\K}e^{-f(x)}\rmd x
+\int_{\ball_{2}(0,R)\cap \K^c}e^{-U^{B,\lambda}(x)}\rmd x
+\int_{\ball_{2}(0,R)^c}e^{-U^{B,\lambda}(x)}\rmd x \,.
\end{align*}
We now derive the bound for the second and the third terms on the right-hand side of the previous display.
Since $f\geqslant l$, we then find
\begin{align*}
\int_{\ball_{2}(0,R)\cap \K^c}e^{-U^{B,\lambda}(x)}\rmd x
\leqslant e^{-l}\int_{\ball_{2}(0,R)}\rmd x\,.
\end{align*}
Thus,
\begin{align*}
\int_{\ball_{2}(0,R)\cap \K^c}e^{-U^{B,\lambda}(x)}\rmd x
\leqslant e^{-l}\int_{\ball_{2}(0,R)}\rmd x
=e^{-l} \frac{\pi^{p/2}}{\Gamma(\frac{d}{2}+1)}R_{}^p
<\infty
\,.
\end{align*}
It remains to show $\int_{\rp}e^{-U^{B,\lambda}(s)}\rmd s<\infty$. 
\begin{align*}
    \int_{\ball_2(0,R)^c}e^{-U^{B,\lambda}(x)}\rmd x&=\int_{\ball_2(0,R)^c}e^{-f(x)}\exp\left(-\inf_{x'\in D^{-1}\K}\dfrac{1}{2\lambda^2}\|D^{-1}x-x'\|_2^2\right)\rmd x\\
    &\leqslant e^{-l}\int_{\ball_2(0,R)^c}\exp\left(-\inf_{x'\in \K}\dfrac{1}{\lambda^2\|D\|_2^2}\|x-x'\|_2^2\right)\rmd x\\
    &\leqslant e^{-l}\int_{\ball_2(0,R)^c}\exp\left(-\dfrac{1}{\lambda^2\|Q^{-1}\|_2}(\|x\|_2-R)^2\right).
\end{align*}
By Fubini's theorem and the fact that
\begin{align*}
e^{-\frac{1}{\alpha}(\,\normtwo{x}-R)^2}
=\int_0^\infty \frac{2t}{\alpha} e^{-\frac{t^2}{\alpha}}\1_{[\,\normtwo{x}-R,\infty)}(t)\rmd t\,,
\end{align*}
we obtain
\begin{align*}
\int_{\ball_{2}(0,R)^c}e^{-U^{B,\lambda}(x)}\rmd x
\leqslant & e^{-l}\int_0^\infty  \frac{2t}{\lambda^2\|Q^{-1}\|_2} e^{-\frac{t}{\lambda^2\|Q^{-1}\|_2}} 
\int_{x:\,\normtwo{x}\geqslant R}
\1_{[\,\normtwo{x}-R,\infty)}(t) \rmd x \rmd t \\
\leqslant & e^{-l}\int_0^\infty \frac{2t}{\lambda^2\|Q^{-1}\|_2} e^{-\frac{t}{\lambda^2\|Q^{-1}\|_2}} \int_{\ball_{2}(0,R+t)} \rmd x \rmd t \\
= &  e^{-l}\int_0^\infty \frac{2t}{\lambda^2\|Q^{-1}\|_2} e^{-\frac{t}{\lambda^2\|Q^{-1}\|_2}} \frac{\pi^{p/2}}{\Gamma(\frac{p}{2}+1)}({R+t})^d \rmd t\\
<& \infty\,.
\end{align*}
Collecting pieces gives the desired result.

We now consider the Gauge projection.\\
\textbf{II. Gauge projection}\\
\noindent\textbf{Strong Convexity.} 
Since $\K$ is convex, its associated Gauge function $g_{\K}$ is convex. Moreover, the function $h(u) = (u - 1)^2$ is convex and non-decreasing on $[1, \infty)$. Therefore, the composition $d_{\K} := h \circ g_{\K}$ is convex. The strong convexity of $U^\lambda$ follows from the strong convexity of $f$.

\noindent\textbf{Smoothness. } Denote the original Gauge function by $\gamma_{\K}(x):=\inf\{t\geqslant 0:x\in t\K\}$, therefore $g_{\K}=1$ inside $\K$ and $g_{\K}(x)=\gamma_{\K}(x)$ outside $\K$. For $d_{\K}(x)=(g_{\K}(x)-1)^2$, we analyze its smoothness outside $\K$.
In $\K^c$, the Hessian matrix is
\begin{align*}
    H_{d_{\K}}=2\nabla g_{\K}(\nabla g_{\K})^\top+2(g_{\K}-1)H_{g_{\K}}\,,
\end{align*}
where $H_{g_{\K}}=H_{\gamma_{\K}}$ since $g_{\K}=\gamma_{\K}$ in this region, therefore $\normtwo{\nabla g_{\K}(x)}=\normtwo{\nabla\gamma_{\K}(x)}$. For any direction $h\in\rp$, the directional derivative satisfies $\gamma_{\K}(x+h)-\gamma_{\K}(x)\geqslant\nabla\gamma_{\K}(x)\cdot h$ by convexity of $\gamma_{\K}$. Simultaneously, sublinearity implies $\gamma_{\K}(x+h)\leqslant\gamma_{\K}(x)+\gamma_{\K}(h)$. Combining these we have $\nabla\gamma_{\K}(x)\cdot h\leqslant\gamma_{\K}(h)$. Choose $h=\nabla\gamma_{\K}(x)/\normtwo{\nabla\gamma_{\K}(x)}$, it then holds that $\normtwo{\nabla\gamma_{\K}(x)}\leqslant\gamma_{\K}(h)\leqslant\frac{1}{r}$. The first term $2\nabla g_{\K}(\nabla g_{\K})^\top$ is rank-1, with its spectral norm bounded by $2\|\nabla\gamma_{\K}\|_2^2\leqslant 2/r^2$. For the second term, on the unit sphere $\mathbb S^{p-1}$, the continuity of $H_{\gamma_{\K}}$ over the compact $\K$ ensures a finite maximum eigenvalue $C_{\K}$, where $C_{\K}$ only depends on $\K$. By 1-homogeneity of $\gamma_{\K}$, we have $H_{\gamma_{\K}}(tx)=\frac{1}{t}H_{\gamma_{\K}}(x)$, it implies
\begin{align*}
    \normtwo{H_{g_{\K}}(x)}=\normtwo{H_{\gamma_{\K}}(x)}\leqslant\dfrac{C_{\K}}{\normtwo{x}}\,.
\end{align*}
We also have $g_{\K}(x)\leqslant\normtwo{x}/r$. Combining these, the eigenvalues of $H_{d_{\K}}$ in $\K^c$ are uniformly bounded by $2/r^2+2C_{\K}/r$. Consequently, $d_{\K}$ is $\frac{2(1+C_{\K}r)}{r^2}$-smooth in $\K^c$, and $d_{\K}\equiv 0$ implies $0$-smoothness in $\K$. For the global smoothness over $\rp$, observe that for any $x\in\K$ and $y\in\K^c$, let $z$ be the intersection of $\partial\K$ with line segment $\vec{xy}$, $\normtwo{\nabla d_{\K}(y)-\nabla d_{\K}(x)}=\normtwo{\nabla d_{\K}(y)}=\normtwo{\nabla d_{\K}(y)-\nabla d_{\K}(z)}$ by the fact that $\nabla d_{\K}\big|_{\partial\K}\equiv 0$, then
\begin{align*}
    \normtwo{\nabla d_{\K}(y)-\nabla d_{\K}(x)}=\normtwo{\nabla d_{\K}(y)-\nabla d_{\K}(z)}\leqslant \dfrac{2(1+C_{\K}r)}{r^2}\normtwo{y-z}\leqslant\dfrac{2(1+C_{\K}r)}{r^2}\normtwo{y-x}
\end{align*}
Thus, $d_{\K}$ is globally $\frac{2(1+C_{\K}r)}{r^2}$-smooth.

To show that $\int_{\rp} e^{-U^{G,\lambda}(x)}\rmd x<\infty.$ 
Note that
\begin{align*}
\int_{\rp}e^{-U^{G,\lambda}(x)}\rmd x
=&\int_{\K}e^{-f(x)}\rmd x
+\int_{\ball_{2}(0,R)\cap \K^c}e^{-U^{G,\lambda}(x)}\rmd x
+\int_{\ball_{2}(0,R)^c}e^{-U^{G,\lambda}(x)}\rmd x \,.
\end{align*}
The first term and the second term have the same bound as in the proof of the Bregman projection case, since we only need the property that $\ell_{\K}^{G,\lambda}\geqslant 0$. For the third term, similarly,
by Fubini's theorem and the fact that
\begin{align*}
e^{-\frac{1}{2\lambda^2}(g_{\K}(x)-1)^2}
=\int_0^\infty \frac{t}{\lambda^2} e^{-\frac{t^2}{2\lambda^2}}\1_{[g_{\K}(x)-1,\infty)}(t)\rmd t\,,
\end{align*}
we obtain
\begin{align*}
\int_{\ball_{2}(0,R)^c}e^{-U^{G,\lambda}(x)}\rmd x
&\leqslant e^{-l}\int_0^\infty  \frac{t}{\lambda^2} e^{-\frac{t}{2\lambda^2}} 
\int_{x:\,\normtwo{x}\geqslant R}
\1_{[g_{\K}(x)-1,\infty)}(t) \rmd x \rmd t \\
&\leqslant e^{-l}\int_0^\infty \frac{t}{\lambda^2} e^{-\frac{t}{2\lambda^2}} \int_{\ball_{2}(0,(1+t)R)} \rmd x \rmd t \\
&= e^{-l}\int_0^\infty \frac{t}{\lambda^2} e^{-\frac{t}{2\lambda^2}} \frac{\pi^{p/2}}{\Gamma(\frac{p}{2}+1)}({(1+t)R})^p \rmd t\\
&=\dfrac{e^{-l}\pi^{p/2}R^p}{\Gamma(p/2+1)}\int_0^\infty se^{-s^2/2}(1+\lambda s)^p\rmd s\\
&< \infty\,.
\end{align*}
% Collecting pieces gives the desired result.
Then we reach the desired result.
\end{proof}

\subsection{Proof of Theorem~\ref{thm:rlmc}}

  Set $\kappa=M^\lambda/m.$
  The triangle inequality for the Wasserstein distance provides us with 
\begin{align}
\label{eq:ws2}
\wstwo(\nu_n^{\sf CRLMC},\nu) \leqslant \wstwo(\nu_n^{\sf CRLMC},\nu^\lambda)+\wstwo(\nu^\lambda,\nu)\,.
\end{align}
By Theorem 1 in~\cite{yu2023langevin}, it holds that
\begin{align}
\label{eq:ws2rlmc}
\wstwo(\nu_n^{\sf CRLMC},\nu^\lambda)&\leqslant 1.1e^{-\frac{mnh}{2}}\wstwo(\nu_0^{\sf CRLMC},\nu^\lambda)+(2.4\sqrt{\kappa M^{\lambda}h}+1.77)M^{\lambda}h\sqrt{p/m}
\end{align}
provided that $M^\lambda h+\sqrt{\kappa}(M^\lambda h)^{3/2}\leqslant 1/4.$ Using the triangle inequality for the Wasserstein distance again gives
\begin{align*}
    \wstwo(\nu_n^{\sf CRLMC},\nu)\leqslant 1.1e^{-\frac{mnh}{2}}\wstwo(\nu_0^{\sf CRLMC},\nu)+(2.4\sqrt{\kappa M^{\lambda}h}+1.77)M^{\lambda}h\sqrt{p/m}+2.1\wstwo(\nu^\lambda,\nu)
\end{align*}
The first claim follows readily from Proposition~\ref{prop:w2_tight}.\\
We then proceed to prove the second claim. By the triangle inequality and the monotonicity of Wasserstein distance, we have
\begin{align*}
\wsone(\nu_n^{\sf CRLMC},\nu)
&\leqslant
\wsone(\nu_n^{\sf CRLMC},\nu^{\lambda})
+\wsone(\nu^{\lambda},\nu)\\
&\leqslant \wstwo(\nu_n^{\sf CRLMC},\nu^{\lambda})
+\wsone(\nu^{\lambda},\nu)\,.
\end{align*}
When $\nu_0^{\sf CRLMC}$ is the Dirac mass at the minimizer of the function $U^\lambda$,  combining this with previous display and display~\eqref{eq:ws2rlmc} gives
\begin{align*}
\wsone(\nu_n^{\sf CRLMC},\nu)
&\leqslant 1.1e^{-\frac{mnh}{2}}\wstwo(\nu_0^{\sf CRLMC},\nu^\lambda)+(2.4\sqrt{\kappa M^{\lambda}h}+1.77)M^{\lambda}h\sqrt{p/m}
+ \wsone(\nu^{\lambda},\nu)\\
&\leqslant  1.1e^{-\frac{mnh}{2}} \sqrt{p/m}+(2.4\sqrt{\kappa M^{\lambda}h}+1.77)M^{\lambda}h\sqrt{p/m}+\wsone(\nu^{\lambda},\nu)\,.
\end{align*}
Invoking Proposition~\ref{prop:w2_tight} yields the desired result.

\subsection{Proof of Corollary~\ref{cor:rlmc}}

Note that
\begin{align*}
    \wstwo(\nu_n^{\sf CRLMC},\nu)\leqslant 1.1e^{-\frac{mnh}{2}}\sqrt{p/m}+(2.4\sqrt{\kappa M^{\lambda}h}+1.77)M^{\lambda}h\sqrt{p/m}+2.1C(p,2)\lambda^{\frac{1}{2}+\frac{1}{p}}
\end{align*}
To reach the best convergence rate, we choose
\[
\lambda=\left(\dfrac{19.2M_0^2}{2.1m}\right)^{\frac{2p}{9p+2}}\dfrac{p^{3p/(9p+2)}}{(C(p,2)(p+2))^{2p/(9p+2)}}h^{\frac{3p}{9p+2}}\,.
\]
Substituting $\lambda$ into the upper bound, we see that to attain accuracy $\varepsilon$ it suffices to choose
\[
h\leqslant A_1\varepsilon^{\frac{18p+4}{3p+6}}\dfrac{p^{8p/(3p+6)}}{C(p,2)^{16p/(3p+6)}}M_0^{-\frac{4}{3}}m^{-\frac{7p-2}{3p+6}}
\]
where $A_1=[3\times(2.1\times(9.6/2.1)^{(p+2)/(9p+2)}+2.4\times(19.2/2.1)^{-8p/(9p+2)})]^{-(18p+4)/(3p+6)}.$
Then
\begin{align*}
    n\geqslant \frac{2}{mh}\ln\big(\frac{3.3}{\varepsilon}\big)\,.
\end{align*}
We now consider the case of $q=1$.
Note that 
\begin{align*}
    \wsone(\nu_n^{\sf CRLMC},\nu)\leqslant 1.1e^{-\frac{mnh}{2}}\sqrt{p/m}+(2.4\sqrt{\kappa M^{\lambda}h}+1.77)M^{\lambda}h\sqrt{p/m}+C(p,1)\lambda\,.
\end{align*}
Then, the optimal $\lambda$ is
\[
\lambda=9.6^{1/5}p^{1/10}C(p,1)^{-1/5}M_0^{2/5}m^{-1/5} h^{3/10}\,.
\]
Therefore, to achieve the $\varepsilon$ accuracy, we choose 
\[
h\leqslant 0.0027\,\varepsilon^{\frac{10}{3}}\dfrac{p^{4 /3}}{C(p,1)^{8/3}}M_0^{-\frac{4}{3}}m^{-1}\,,
\]
and 
\[
n\geqslant\dfrac{2}{mh}\ln\left(\dfrac{3.3}{\varepsilon}\right)\,.
\]

\subsection{Proof of Theorem~\ref{thm:rklmc}}

    The proof employs a similar technique to that used in the proof of Theorem~\ref{thm:rlmc}. By the triangle inequality and Theorem 2 from \cite{yu2023langevin}, we have
    \begin{align*}
        \wstwo(\nu_n^{\sf CRKLMC},\nu)&\leqslant \wstwo(\nu^\lambda,\nu)+\wstwo(\nu_n^{\sf CRKLMC},\nu^\lambda)\\
        &\leqslant 1.6e^{-mnh}\wstwo(\nu_0^{\sf CRKLMC},\nu^\lambda)+0.2(\gamma h)^3\sqrt{\kappa p/m}+10(\gamma h)^{3/2}\sqrt{p/m}+\wstwo(\nu^\lambda,\nu)\\
        &\leqslant 1.6e^{-mnh}\wstwo(\nu_0^{\sf CRKLMC},\nu)+0.2(\gamma h)^3\sqrt{\kappa p/m}+10(\gamma h)^{3/2}\sqrt{p/m}+2.6\wstwo(\nu^\lambda,\nu)\,.
    \end{align*}
    The first claim follows from Proposition~\ref{prop:w2_tight}.
    For the second claim, it holds that
    \begin{align*}
        \wsone(\nu_n^{\sf CRKLMC},\nu)&\leqslant \wsone(\nu^\lambda,\nu)+\wstwo(\nu_n^{\sf CRKLMC},\nu^\lambda)\\
        &\leqslant 1.6e^{-mnh}\wstwo(\nu_0^{\sf CRKLMC},\nu^\lambda)+0.2(\gamma h)^3\sqrt{\kappa p/m}+10(\gamma h)^{3/2}\sqrt{p/m}+\wsone(\nu^\lambda,\nu)\,.
      \end{align*}
Applying the same technique used in the proof of Theorem \ref{thm:rlmc} yields the desired result.

\subsection{Proof of Corollary~\ref{cor:rklmc}}

Note that 
\begin{align*}
    \wstwo(\nu_n^{\sf CRKLMC},\nu)\leqslant 1.6e^{-mnh}\sqrt{p/m}+0.2(\gamma h)^3\sqrt{\kappa p/m}+10(\gamma h)^{3/2}\sqrt{p/m}+2.6C(p)\lambda^{\frac{1}{2}+\frac{1}{p}}\,.
\end{align*}
To achieve the $\varepsilon$ accuracy, we choose the optimal $\lambda$, the corresponding step size
 and the sample size as follows
 \begin{align*}
    \lambda=\left(\dfrac{150M_0^{7/2}}{m}\right)^{\frac{2p}{15p+2}}\dfrac{p^{p/(15p+2)}}{C(p,2)^{2p/(15p+2)}}h^{\frac{6p}{15p+2}}\,.
\end{align*}
\begin{align*}
    h\leqslant A_2\varepsilon^{\frac{15p+2}{3p+6}}\dfrac{p^{7p/(3p+6)}}{C(p,2)^{14p/(3p+6)}}M_0^{-\frac{7}{6}}m^{-\frac{13p-2}{6p+12}}
\end{align*}
where $A_2=[3\times(25\times 150^{-14p/(15p+2)}+2.6\times 75^{(p+2)/(15p+2)})]^{-(15p+2)/(3p+6)}.$
\begin{align*}
    n\geqslant \dfrac{1}{mh}\ln\left(\dfrac{4.8}{\varepsilon}\right)
\end{align*}
Using the same trick as in the proof of Corollary~\ref{cor:rklmc}, for $q=1$, we have
\begin{align*}
    \wsone(\nu_n^{\sf CRKLMC},\nu)\leqslant 1.6e^{-mnh}\sqrt{p/m}+0.2(\gamma h)^3\sqrt{\kappa p/m}+10(\gamma h)^{3/2}\sqrt{p/m}+C(p,1)\lambda\,.
\end{align*}
Then, the optimal $\lambda$ is
\begin{align*}
    \lambda=1.91h^{3/8}p^{1/16}C(p,1)^{-1/8}M_0^{7/16}m^{-1/8}\,.
\end{align*}
Thus, we choose 
\begin{align*}
    h\leqslant 0.0067\,\varepsilon^{\frac{8}{3}}\dfrac{p^{7/6}}{C(p,1)^{7/3}}M_0^{-7/6}m^{-1}\,,
\end{align*}
and
\begin{align*}
    n\geqslant\dfrac{1}{mh}\ln\left(\dfrac{5}{\varepsilon}\right)\,.
\end{align*}

\section{Proofs of Sections~\ref{sec:lmc} and~\ref{sec:klmc}}

\subsection{Proof of Theorem~\ref{thm:lmc}}
% \begin{proof}
    By Theorem 9 in~\cite{durmus2019analysis} and the triangle inequality for the Wasserstein distance, it holds that
    \begin{align*}
        \wstwo(\nu_n^{\sf CLMC},\nu) &\leqslant \wstwo(\nu_n^{\sf CLMC},\nu^\lambda)+\wstwo(\nu^\lambda,\nu)\\
        &\leqslant e^{-{mnh}}\wstwo(\nu_0^{\sf CLMC},\nu^{\lambda})+ \sqrt{\frac{2M^\lambda ph}{m}}+\wstwo(\nu^\lambda,\nu)
    \end{align*}
provided that $h\leqslant 1/M^\lambda.$ 
For $q=1$, we have
\begin{align*}
    \wsone(\nu_n^{\sf CLMC},\nu)\leqslant e^{-mnh}\sqrt{p/m}+\sqrt{\frac{2M^\lambda ph}{m}}+\wsone(\nu^\lambda,\nu)
\end{align*}
as desired.
Invoking Proposition~\ref{prop:w2_tight} gives the desired result.

\subsection{Proof of Corollary~\ref{cor:lmc}}

Note that
\begin{align*}
    \wstwo(\nu_n^{\sf CLMC},\nu)\leqslant e^{-mhn} \sqrt{p/m}+ \sqrt{\frac{2M^\lambda ph}{m}}+2C(p,2)\lambda^{\frac{1}{2}+\frac{1}{p}}\,.
\end{align*}
The optimal $\lambda$ is 
\begin{align*}
    \lambda=\left(\dfrac{2M_0}{m}\right)^{\frac{p}{3p+2}}\dfrac{p^{3p/(3p+2)}}{(C(p,2)(p+2))^{2p/(3p+2)}}h^{\frac{p}{3p+2}}
\end{align*}
To achieve the $\varepsilon$ accuracy, we choose 
\begin{align*}
    h\leqslant A_3\varepsilon^{\frac{6p+4}{p+2}}\dfrac{p^{2p/(p+2)}}{C(p,2)^{4p/(p+2)}}M_0^{-1}m^{-\frac{2p}{p+2}}\,
\end{align*}
where $A_3=[3\times(2^{(p+2)/(6p+4)}+2^{(5p+2)/(6p+4)})]^{-(6p+4)/(p+2)}$, and
\begin{align*}
    n\geqslant \dfrac{2}{mh}\ln\left(\dfrac{3}{\varepsilon}\right)\,.
\end{align*}
For $q=1$, we have
\begin{align*}
    \wsone(\nu_n^{\sf CLMC},\nu)\leqslant e^{-mnh}\sqrt{p/m} +\sqrt{\frac{2M^\lambda ph}{m}}+C(p,1)\lambda\,.
\end{align*}
Then, the optimal $\lambda=1.19h^{1/4}p^{1/4}C(p,1)^{-1/2}M_0^{1/4}m^{-1/4}$, thus we choose
\begin{align*}
    h\leqslant 0.00039\,\varepsilon^{4}\dfrac{p}{C(p,1)^2}M_0^{-1}m^{-1}\,,
\end{align*}
and
\begin{align*}
    n\geqslant \dfrac{2}{mh}\ln\left(\dfrac{3}{\varepsilon}\right)\,.
\end{align*}

\subsection{Proof of Theorem~\ref{thm:klmc}}

By Theorem 3 from~\cite{yu2023langevin}, following the same approach in the proof of Theorems~\ref{thm:rlmc}-~\ref{thm:lmc}, we arrive at the desired results.

\subsection{Proof of Corollary~\ref{cor:klmc}}

Note that
\begin{align*}
    \wstwo(\nu_n^{\sf CKLMC},\nu)&\leqslant 2e^{-mhn}\sqrt{p/m}+0.9\gamma h\sqrt{\kappa p/m}+3C(p,2)\lambda^{\frac{1}{2}+\frac{1}{p}}\,,
\end{align*}
Then the optimal $\lambda$ is chosen to be
\begin{align*}
    \lambda=\left(\dfrac{9M_0^{3/2}}{m}\right)^{\frac{2p}{7p+2}}\dfrac{p^{3p/(7p+2)}}{(C(p,2)(p+2))^{2p/(7p+2)}}h^{\frac{2p}{7p+2}}\,.
\end{align*}
Thus, to achieve the $\varepsilon$ accuracy, it is enough to let
\begin{align*}
    h\leqslant A_4\varepsilon^{\frac{7p+2}{p+2}}\dfrac{p^{3p/(p+2)}}{C(p,2)^{6p/(p+2)}}M_0^{-3/2}m^{-\frac{5p-2}{2p+4}}\,,
\end{align*}
and
\begin{align*}
    n\geqslant \dfrac{1}{mh}\ln\left(\dfrac{6}{\varepsilon}\right)\,.
\end{align*}
For $q=1$, it holds that
\begin{align*}
    \wsone(\nu_n^{\sf CKLMC},\nu)\leqslant 2e^{-mhn} \sqrt{p/m} +0.9\gamma h\sqrt{\kappa p/m}+C(p,1)\lambda\,.
\end{align*}
The optimal $\lambda=1.92h^{1/4}p^{1/8}C(p,1)^{-1/4}M_0^{3/8}m^{-1/4}$, then we choose
\begin{align*}
    h\leqslant 0.00029\varepsilon^4\dfrac{p^{3/2}}{C(p,1)^3}M_0^{-3/2}m^{-1}\,,
\end{align*}
and
\begin{align*}
    n\geqslant \dfrac{1}{mh}\ln\left(\dfrac{6}{\varepsilon}\right)\,.
\end{align*}

\section{Numerical Studies}
\label{app:numerical}
\begin{figure}[H]
\centering
\includegraphics[width=1\linewidth]{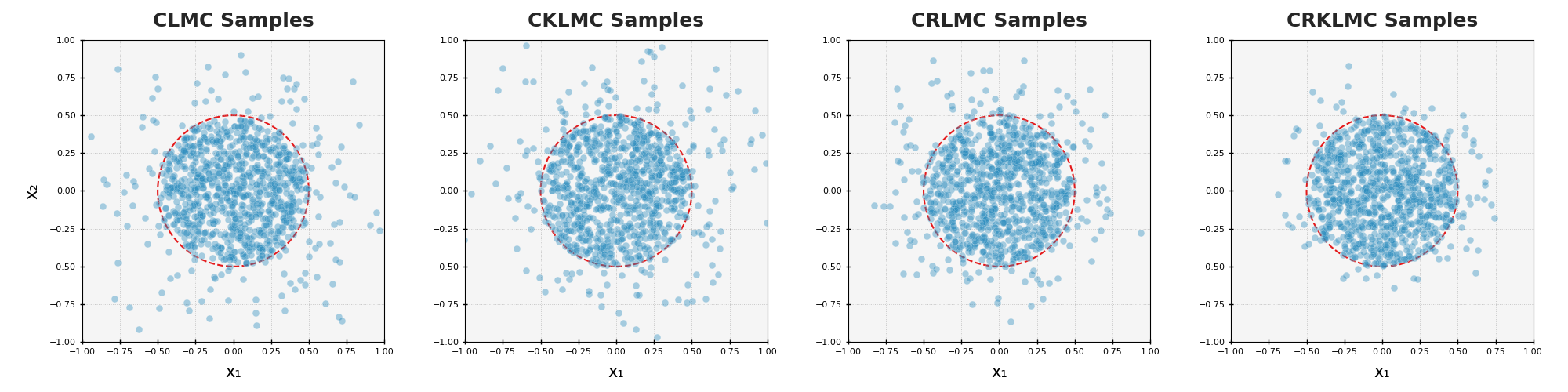}
\caption{Scatter plots of samples generated by $\{\text{CL,CKL,CRL,CRKL}\}\text{MC}$ algorithms.}
\label{fig:radial_more}
\end{figure}

\begin{figure}[H]
\centering
\includegraphics[width=1\linewidth]{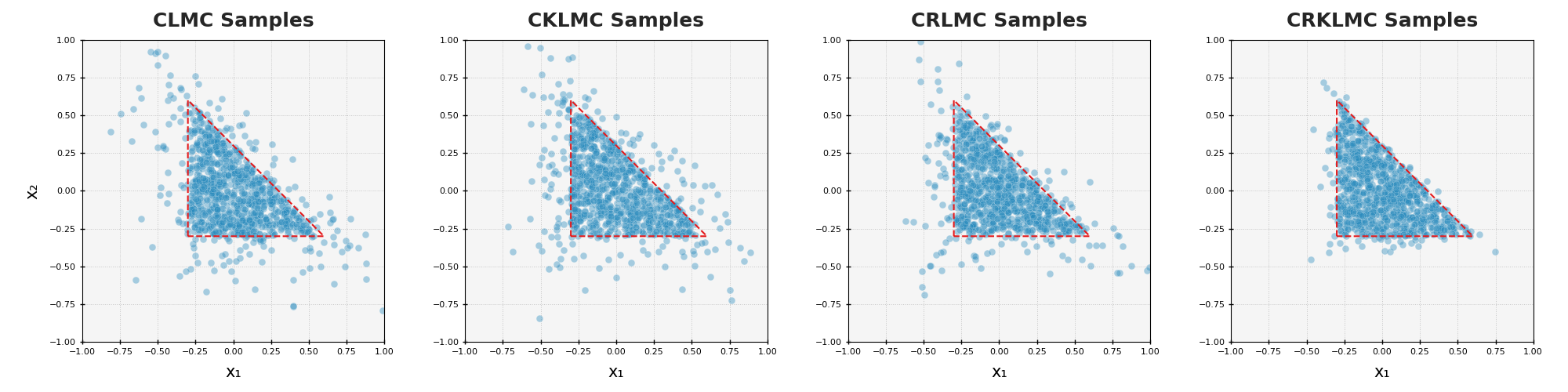}
\caption{Scatter plots of samples generated by $\{\text{CL,CKL,CRL,CRKL}\}\text{MC}$ algorithms.}
\label{fig:tria_more}
\end{figure}

Figures~\ref{fig:radial_more} and~\ref{fig:tria_more} show scatter plots of the four algorithms with $n = 2000$ and $N = 1000$. The red dashed lines indicate the boundaries of the constraint sets.
These plots exhibit the same trends as those presented in the main text: CRKLMC demonstrates the best convergence, with samples most tightly concentrated within the constraint set. CRLMC outperforms CLMC, and CRKLMC improves upon CKLMC, illustrating the advantage of midpoint randomization over Euler discretization.

\end{document}